\documentclass[10pt,twocolumn,letterpaper]{article}

\usepackage{style}
\usepackage{times}
\usepackage{epsfig}
\usepackage{graphicx}
\usepackage{amsmath}
\usepackage{amssymb}
\usepackage{mathtools}
\usepackage{subcaption}
\usepackage{comment}
\usepackage{mathtools}
\usepackage{acronym}
\usepackage{caption}
\usepackage{array}
\usepackage{tabularx}
\usepackage{subcaption}
\usepackage{booktabs}
\usepackage{multicol}
\usepackage{multirow}
\usepackage[norelsize, linesnumbered, ruled, lined, boxed, commentsnumbered]{algorithm2e}

\usepackage[pagebackref=true,breaklinks=true,letterpaper=true,colorlinks,bookmarks=false]{hyperref}

\cvprfinalcopy

\ifcvprfinal\fi

\DeclareMathOperator*{\argmin}{arg\,min}

\begin{document}

\title{ Image2StyleGAN++: How to Edit the Embedded Images?}

\author{Rameen Abdal\\ KAUST \\{\tt\small rameen.abdal@kaust.edu.sa} \and Yipeng Qin \\ Cardiff University \\ {\tt\small qiny16@cardiff.ac.uk} \and Peter Wonka\\ KAUST \\{\tt\small pwonka@gmail.com}}
\twocolumn[{%
\renewcommand\twocolumn[1][]{#1}%
\maketitle
\begin{center}

     \includegraphics[width= 0.99\linewidth]{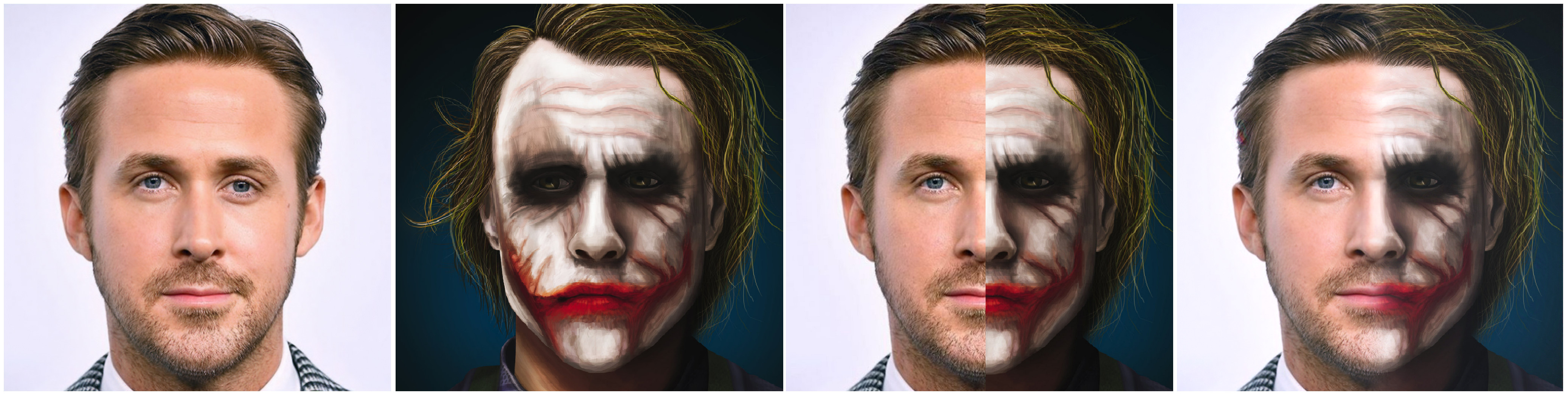}
          \hspace*{0.0cm}(a)\hspace{4.0cm}(b)\hspace{4.0cm}(c)\hspace{4.0cm}(d)

    \captionof{figure}{(a) and (b): input images; (c): the ``two-face'' generated by naively copying the left half from (a) and the right half from (b); (d): the ``two-face'' generated by our Image2StyleGAN++ framework.}
    \label{fig:teaser}
\end{center}
}]

\begin{abstract}

We propose Image2StyleGAN++, a flexible image editing framework with many applications.
Our framework extends the recent Image2StyleGAN~\cite{abdal2019image2stylegan} in three ways. 
First, we introduce noise optimization as a complement to the $W^+$ latent space embedding. Our noise optimization can restore high frequency features in images and thus significantly improves the quality of reconstructed images, \eg a big increase of PSNR from 20 dB to 45 dB.
Second, we extend the global $W^+$ latent space embedding to enable local embeddings.
Third, we combine embedding with activation tensor manipulation to perform high quality local edits along with global semantic edits on images.
Such edits motivate various high quality image editing applications, \eg image reconstruction, image inpainting, image crossover, local style transfer, image editing using scribbles, and attribute level feature transfer.
Examples of the edited images are shown across the paper for visual inspection.
\end{abstract}

\section{Introduction}

Recent GANs~\cite{STYLEGAN2018,BigGAN2019} demonstrated that synthetic images can be generated with very high quality. This motivates research into embedding algorithms that embed a given photograph into a GAN latent space. Such embedding algorithms can be used to analyze the limitations of GANs~\cite{bau2019seeing}, do image inpainting~\cite{demir2018patch,yu2018generative,yu2018freeform,webster2019detecting}, local image editing~\cite{zhu2016generative,Jo_2019_ICCV}, global image transformations such as image morphing and expression transfer~\cite{abdal2019image2stylegan}, and few-shot video generation~\cite{wang2018vid2vid,wang2019fewshot}.

In this paper, we propose to extend a very recent embedding algorithm, Image2StyleGAN~\cite{abdal2019image2stylegan}. In particular, we would like to improve this previous algorithm in three aspects. First, we noticed that the embedding quality can be further improved by including Noise space optimization into the embedding framework. The key insight here is that stable Noise space optimization can only be conducted if the optimization is done sequentially with $W^{+}$ space and not jointly. Second, we would like to improve the capabilities of the embedding algorithm to increase the local control over the embedding. One way to improve local control is to include masks in the embedding algorithm with undefined content. The goal of the embedding algorithm should be to find a plausible embedding for everything outside the mask, while filling in reasonable semantic content in the masked pixels. Similarly, we would like to provide the option of approximate embeddings, where the specified pixel colors are only a guide for the embedding. In this way, we aim to achieve high quality embeddings that can be controlled by user scribbles. In the third technical part of the paper, we investigate the combination of embedding algorithm and direct manipulations of the activation maps (called activation tensors in our paper).

Our main contributions are:
\begin{enumerate}
    \item We propose Noise space optimization to restore the high frequency features in an image that cannot be reproduced by other latent space optimization of GANs. The resulting images are very faithful reconstructions of up to 45 dB compared to about 20 dB (PSNR) for the previously best results.
    \item We propose an extended embedding algorithm into the $W^{+}$ space of StyleGAN that allows for local modifications such as missing regions and locally approximate embeddings.
    \item We investigate the combination of embedding and activation tensor manipulation to perform high quality local edits along with global semantic edits on images.
    \item We apply our novel framework to multiple image editing and manipulation applications. The results show that the method can be successfully used to develop a state-of-the-art image editing software.
\end{enumerate}

\section{Related Work}

\begin{figure*}
        \centering
        \includegraphics[width=\linewidth]{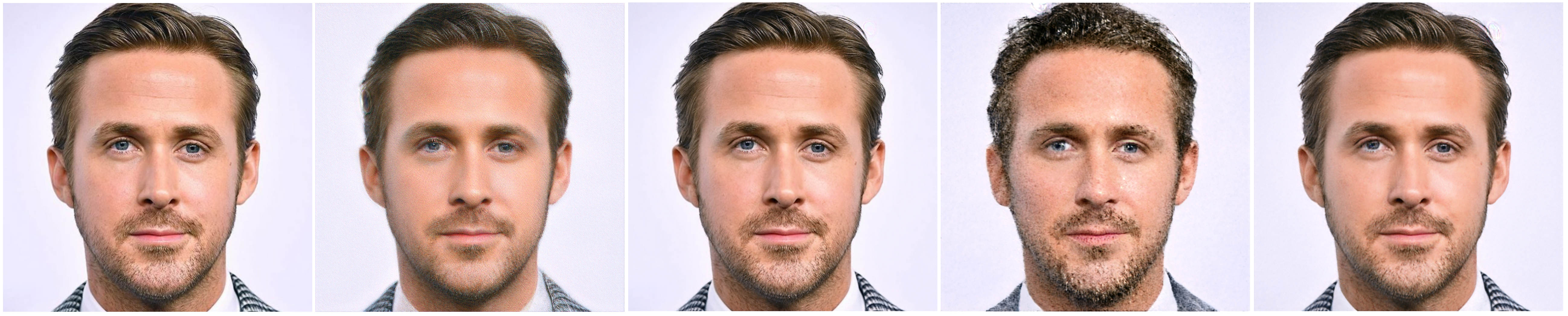}
        \hspace*{0.0cm}(a)\hspace{3.1cm}(b)\hspace{3.1cm}(c)\hspace{3.1cm}(d)\hspace{3.1cm}(e)
        \caption{Joint optimization. (a): target image; (b): image embedded by jointly optimizing $w$ and $n$ using perceptual and pixel-wise MSE loss; (c): image embedded by jointly optimizing $w$ and $n$ using the pixel-wise MSE loss only; (d): the result of the previous column with $n$ resampled; (e): image embedded by jointly optimizing $w$ and $n$ using perceptual and pixel-wise MSE loss for $w$ and  pixel-wise MSE loss for $n$.}
        \label{fig:imp66}
    \end{figure*}
    
     \begin{figure*}
        \centering
        \includegraphics[width=\linewidth]{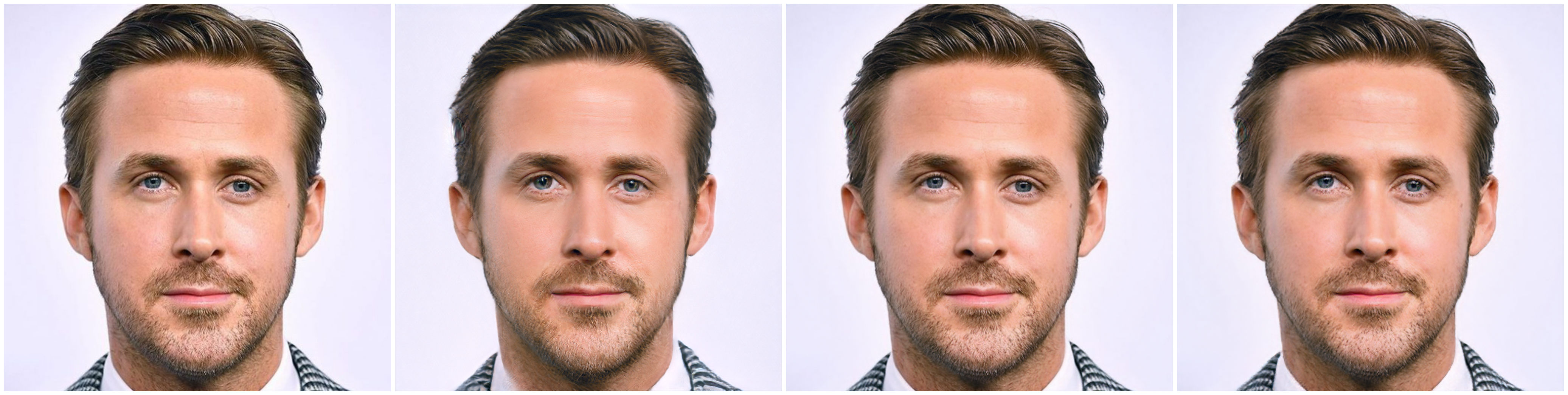}
        \hspace*{0.0cm}(a)\hspace{4.0cm}(b)\hspace{4.0cm}(c)\hspace{4.0cm}(d)
        \caption{Alternating optimization. (a): target image; (b): image embedded by optimizing $w$ only; (c): taking $w$ from the previous column and subsequently optimizing $n$ only; (d): taking the result from the previous column and optimizing $w$ only.}
        \label{fig:imp55}
    \end{figure*}

Generative Adversarial Networks (GANs)~\cite{goodfellow2014generative, DCGAN2015} are one of the most popular generative models that have been successfully applied to many computer vision applications, \eg object detection \cite{ObjectDetection2017},
texture synthesis \cite{TextureSynthesis2016,Xian_2018_CVPR,slossberg2018high},
image-to-image translation \cite{pix2pix2017,CycleGAN2017,park2019SPADE,liu2019few} and video generation \cite{VideoGeneration2016,Tulyakov_2018_CVPR,wang2018vid2vid,wang2019fewshot}. 
Backing these applications are the massive improvements on GANs in terms of architecture~\cite{STYLEGAN2018,BigGAN2019,park2019SPADE,pix2pix2017}, loss function design~\cite{Mao_2017,WGAN2017}, and regularization~\cite{SpectralNormalization2018,gulrajani2017improved}.
On the bright side, such improvements significantly boost the quality of the synthesized images.
To date, the two highest quality GANs are StyleGAN~\cite{STYLEGAN2018} and BigGAN~\cite{BigGAN2019}. 
Between them, StyleGAN produces excellent results for unconditional image synthesis tasks, especially on face images; BigGAN produces the best results for conditional image synthesis tasks (\eg ImageNet~\cite{imagenet_cvpr09}).
While on the dark side, these improvements make the training of GANs more and more expensive that nowadays it is almost a privilege of wealthy institutions to compete for the best performance.
As a result, methods built on pre-trained generators start to attract attention very recently.
In the following, we would like to discuss previous work of two such approaches: embedding images into a GAN latent space and the manipulation of GAN activation tensors.

\paragraph{Latent Space Embedding.}
The embedding of an image into the latent space is a longstanding topic in both machine learning and computer vision.
In general, the embedding can be implemented in two ways: i) passing the input image through an encoder neural network (\eg the Variational Auto-Encoder \cite{VAE2013}); ii) optimizing a random initial latent code to match the input image \cite{Zhu_2016,GANEmbedding2018}.
Between them, the first approach dominated for a long time.
Although it has an inherent problem to generalize beyond the training dataset, it produces higher quality results than the naive latent code optimization methods \cite{Zhu_2016,GANEmbedding2018}.
While recently, Abdal \etal \cite{abdal2019image2stylegan} obtained excellent embedding results by optimizing the latent codes in an enhanced $W^{+}$ latent space instead of the initial $Z$ latent space.
Their method suggests a new direction for various image editing applications and makes the second approach interesting again.

\paragraph{Activation Tensor Manipulation.}
With fixed neural network weights, the expression power of a generator can be fully utilized by manipulating its activation tensors.
Based on this observation, Bau \cite{bau2019gandissect} \etal investigated what a GAN can and cannot generate by locating and manipulating relevant neurons in the activation tensors \cite{bau2019gandissect,bau2019seeing}. 
Built on the understanding of how an object is ``drawn'' by the generator, they further designed a semantic image editing system that can add, remove or change the appearance of an object in an input image \cite{Bau:Ganpaint:2019}.
Concurrently, Fr{\"u}hst{\"u}ck \etal \cite{Fr_hst_ck_2019} investigated the potential of activation tensor manipulation in image blending. Observing that boundary artifacts can be eliminated by by cropping and combining activation tensors at early layers of a generator, they proposed an algorithm to create large-scale texture maps of hundreds of megapixels by combining outputs of GANs trained on a lower resolution.

\section{Overview}

Our paper is structured as follows. First, we describe an extended version of the Image2StyleGAN~\cite{abdal2019image2stylegan} embedding algorithm (See Sec.~\ref{sec:embeds}). We propose two novel modifications:
1) to enable local edits, we integrate various spatial masks into the optimization framework. Spatial masks enable embeddings of incomplete images with missing values and embeddings of images with approximate color values such as user scribbles. In addition to spatial masks, we explore layer masks that restrict the embedding into a set of selected layers. The early layers of StyleGAN~\cite{STYLEGAN2018} encode content and the later layers control the style of the image. By restricting embeddings into a subset of layers we can better control what attributes of a given image are extracted.
2) to further improve the embedding quality, we optimize for an additional group of variables $n$ that control additive noise maps. These noise maps encode high frequency details and enable embedding with very high reconstruction quality.

Second, we explore multiple operations to directly manipulate activation tensors (See Sec.~\ref{ref:activation}). We mainly explore spatial copying, channel-wise copying, and averaging,

Interesting applications can be built by combining multiple embedding steps and direct manipulation steps. As a stepping stone towards building interesting application, we describe in Sec.~\ref{ref:fub} common building blocks that consist of specific settings of the extended optimization algorithm.

Finally, in Sec.~\ref{sec:spp} we outline multiple applications enabled by Image2StyleGAN++: improved image reconstruction, image crossover, image inpainting, local edits using scribbles, local style transfer, and attribute level feature transfer.

\section{An Extended Embedding Algorithm}
\label{sec:embeds}

     \begin{figure}
        \centering
        \includegraphics[width=\linewidth]{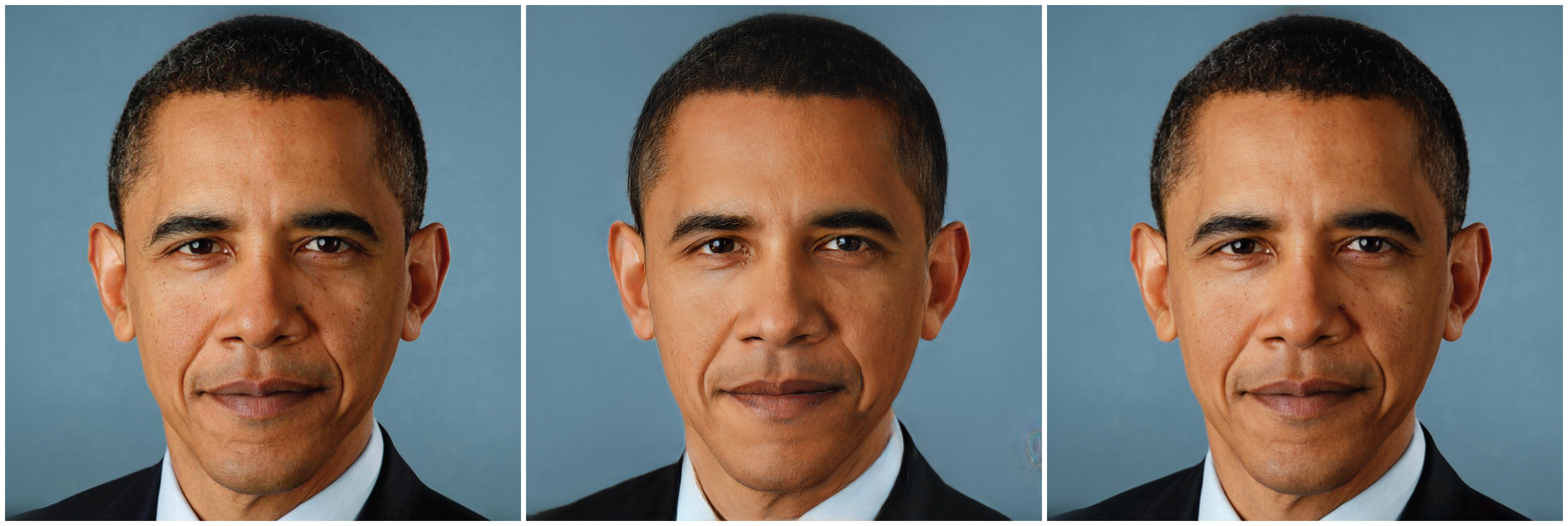}
        \includegraphics[width=\linewidth]{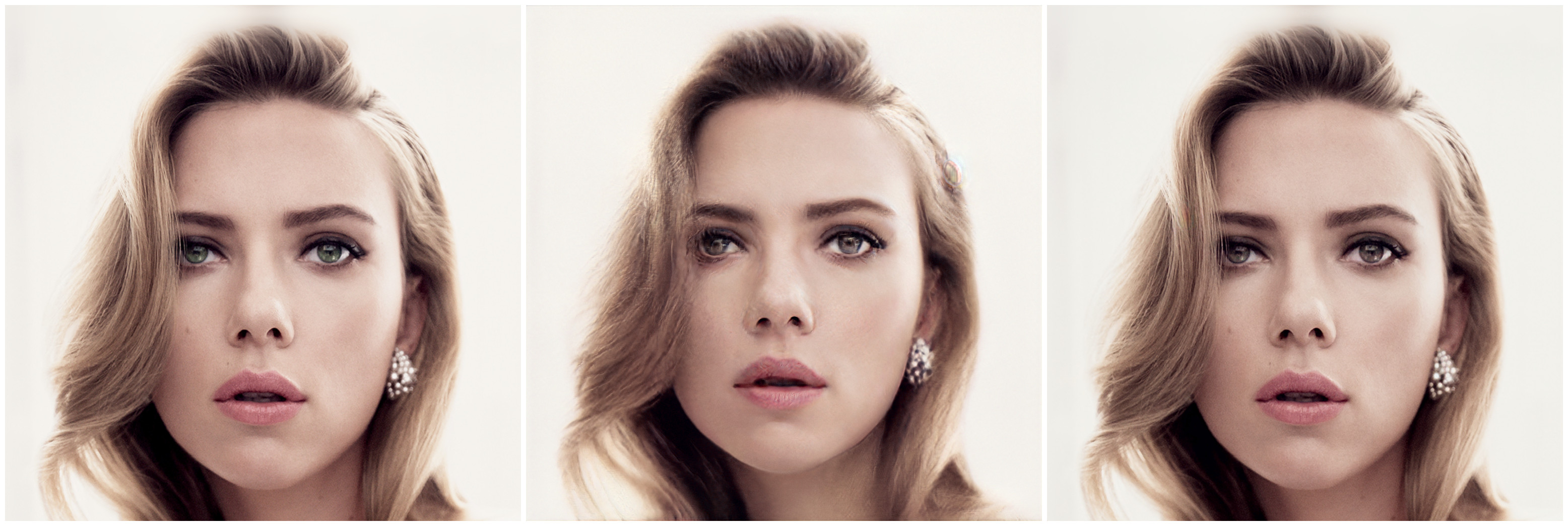}
        \includegraphics[width=\linewidth]{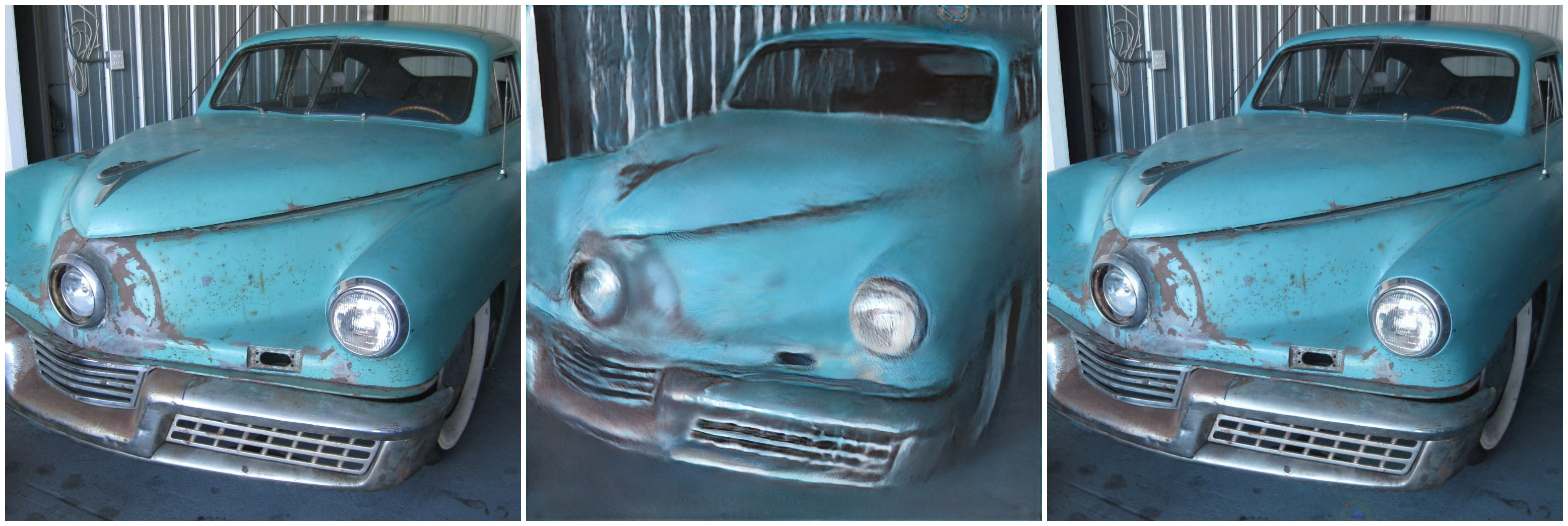}
        \includegraphics[width=\linewidth]{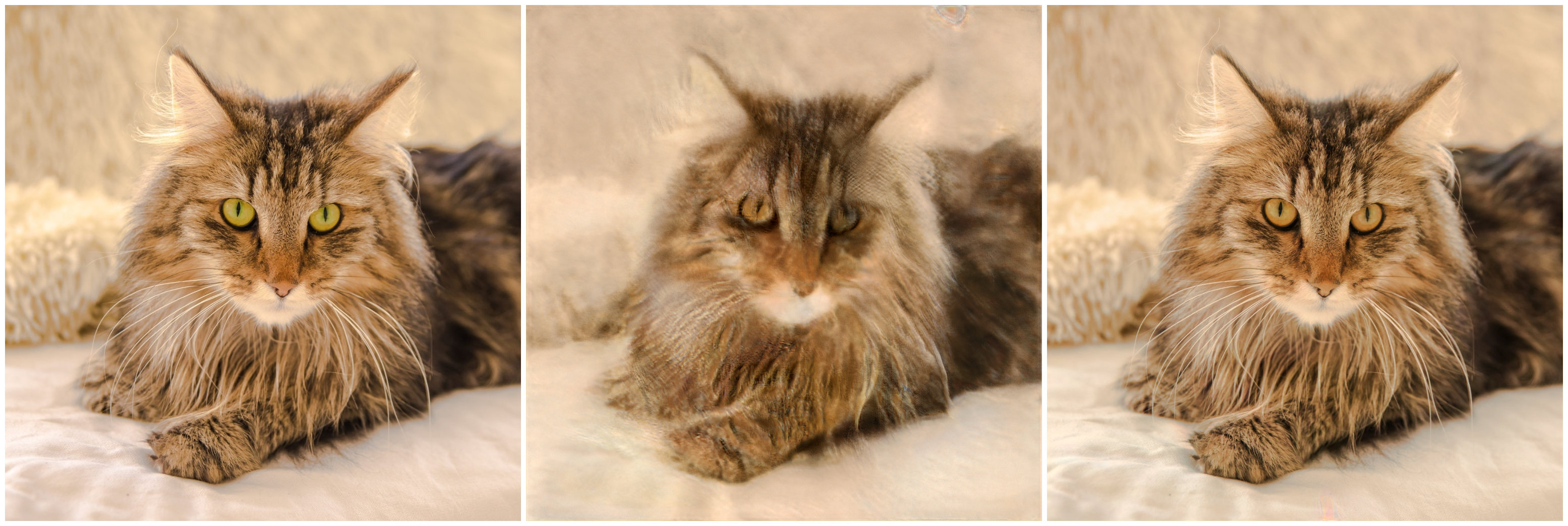}
        
        \caption{First column: original image; Second column: image embedded in $W^{+}$ Space (PSNR 19 to 22 dB); Third column: image embedded in $W^{+}$ and Noise space (PSNR 39 to 45 dB).}
        \label{fig:Latent_noise}
    \end{figure}
We implement our embedding algorithm as a gradient-based optimization that iteratively updates an image starting from some initial latent code.
The embedding is performed into two spaces using two groups of variables; the semantically meaningful $W^{+}$ space and a Noise space $N_{s}$ encoding high frequency details. The corresponding groups of variables we optimize for are $w \in W^{+}$ and $n \in N_{s}$. 
The inputs to the embedding algorithm are target RGB images $x$ and $y$ (they can also be the same image), and up to three spatial masks ($M_s$, $M_m$, and $M_{p}$)

Algorithm \ref{alg:latent_space} is the generic embedding algorithm used in the paper.

\subsection{Objective Function}

Our objective function consists of three different types of loss terms, \ie the pixel-wise MSE loss, the perceptual loss \cite{PerceptualLoss2016,dosovitskiy2016generating}, and the style loss \cite{gatys2016image}.

\begin{equation}
\begin{split}
    L &=  \lambda_s L_{style} (M_s, G(w,n),y) \\ 
    &+ \frac{\lambda_{mse_{1}}}{N} \|M_m \odot (G(w,n) - x)\|_2^2 \\
    &+  \frac{\lambda_{mse_{2}}}{N} \|(1 - M_m) \odot (G(w,n) - y)\|_2^2 \\ 
    &+ \lambda_{p} L_{percept}(M_{p}, G(w, n), x )
\end{split}
\label{eq:loss_function}
\end{equation}
Where $M_s$, $M_m$ , $M_{p}$ denote the spatial masks, $ \odot$ denotes the Hadamard product, $G$ is the StyleGAN generator, $n$ are the Noise space variables, $w$ are the $W^{+}$ space variables, $L_{style}$ denotes style loss from $‘conv3\_3’$ layer of an ImageNet pretrained VGG-16 network~\cite{simonyan2014deep}, $L_{percept}$ is the perceptual loss defined in Image2StyleGAN~\cite{abdal2019image2stylegan}. Here, we use layers $‘conv1\_1’$, $‘conv1\_2’$, $‘conv2\_2’$ and $‘conv3\_3’$ of VGG-16 for the perceptual loss.  Note that the perceptual loss is computed for four layers of the VGG network. Therefore, $M_p$ needs to be downsampled to match the resolutions of the corresponding VGG-16 layers in the computation of the loss function.

\subsection{Optimization Strategies}
Optimization of the variables $w \in W^{+}$ and $n \in N_s$ is not a trivial task. Since only $w \in W^{+}$ encodes semantically meaningful information, we need to ensure that as much information as possible is encoded in $w$ and only high frequency details in the Noise space.

The first possible approach is the joint optimization of both groups of variables $w$ and $n$. Fig.\ref{fig:imp66} (b) shows the result using the perceptual and the pixel-wise MSE loss. We can observe that many details are lost and were replaced with high frequency image artifacts. This is due to the fact that the perceptual loss is incompatible with optimizing noise maps. Therefore, a second approach is to use pixel-wise MSE loss only (see Fig.~\ref{fig:imp66} (c)). Although the reconstruction is almost perfect, the representation $(w,n)$ is not suitable for image editing tasks. In Fig.~\ref{fig:imp66} (d), we show that too much of the image information is stored in the noise layer, by resampling the noise variables $n$. We would expect to obtain another very good, but slightly noisy embedding. Instead, we obtain a very low quality embedding. Also, we show the result of jointly optimizing the variables and using perceptual and pixel-wise MSE loss for $w$ variables and pixel-wise MSE loss for the noise variable. Fig.~\ref{fig:imp66} (e) shows the reconstructed image is not of high perceptual quality. The PSNR score decreases to 33.3 dB. We also tested these optimizations on other images. Based on our results, we do not recommend using joint optimization.

The second strategy is an alternating optimization of the variables $w$ and $n$. In Fig.~\ref{fig:imp55}, we show the result of optimizing $w$ while keeping $n$ fixed and subsequently optimizing $n$ while keeping $w$ fixed. In this way, most of the information is encoded in $w$ which leads to a semantically meaningful embedding.
Performing another iteration of optimizing $w$ (Fig.~\ref{fig:imp55} (d)) reveals a smoothing effect on the image and the PSNR reduces from 39.5 dB to 20 dB. Subsequent Noise space optimization does not improve PSNR of the images. Hence, repetitive alternating optimization does not improve the quality of the image further. In summary, we recommend to use alternating optimization, but each set of variables is only optimized once. First we optimize $w$, then $n$.
\begin{algorithm}[h]
\SetAlgoLined
 \KwIn{images $x,y \in \mathbb{R}^{n \times m \times 3}$; masks $M_s, M_m, M_p$; a pre-trained generator $G(\cdot,\cdot)$; gradient-based optimizer $F'$.}
 \KwOut{the embedded code $(w,n)$}
 Initialize() the code $(w,n)$ = $(w',n')$\;
 \While{not converged}{
  $Loss \leftarrow L(x, y, M_s, M_m, M_p)$\;
  $(w,n) \leftarrow (w,n) - \eta F'(\nabla_{w,n} L,w,n)$\;
 }
 \caption{Semantic and Spatial component embedding in StyleGAN}
 \label{alg:latent_space}
\end{algorithm}

\section{Activation Tensor Manipulations} 
\label{ref:activation}
Due to the progressive architecture of StyleGAN, one can perform meaningful tensor operations at different layers of the network~\cite{Fr_hst_ck_2019,bau2019gandissect}. We consider the following editing operations: spatial copying, averaging, and channel-wise copying. We define activation tensor $A_{l}^{I}$ as the output of the $l$-th layer in the network initialized with variables $(w,n)$ of the embedded image $I$. They are stored as tensors $A_{l}^{I} \in R^{W_l \times H_l \times C_l}$. Given two such tensors $A_{l}^{I}$ and $B_{l}^{I}$, copying replaces high-dimensional pixels $\in R^{1 \times 1 \times C_l}$ in $A_{l}^{I}$ by copying from $B_{l}^{I}$. Averaging forms a linear combination $\lambda A_{l}^{I} + (1-\lambda)B_{l}^{I}$. Channel-wise copying creates a new tensor by copying selected channels from $A_{l}^{I}$ and the remaining channels from $B_{l}^{I}$.
In our tests we found that spatial copying works a bit better than averaging and channel-wise copying.

       \begin{figure}[t]
        \centering

             \includegraphics[width=0.99\linewidth]{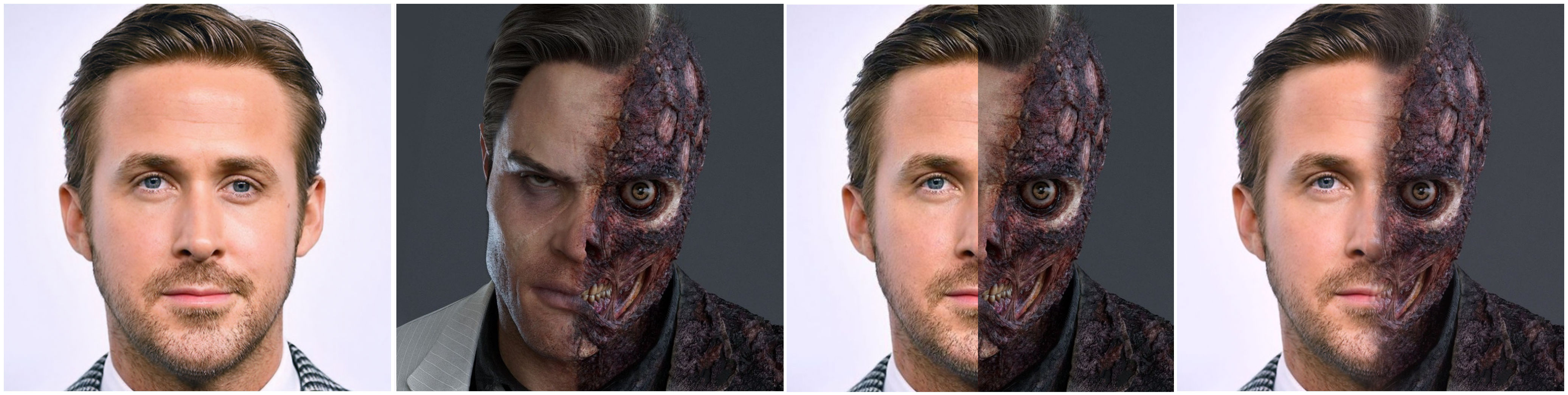}
              \includegraphics[width=0.99\linewidth]{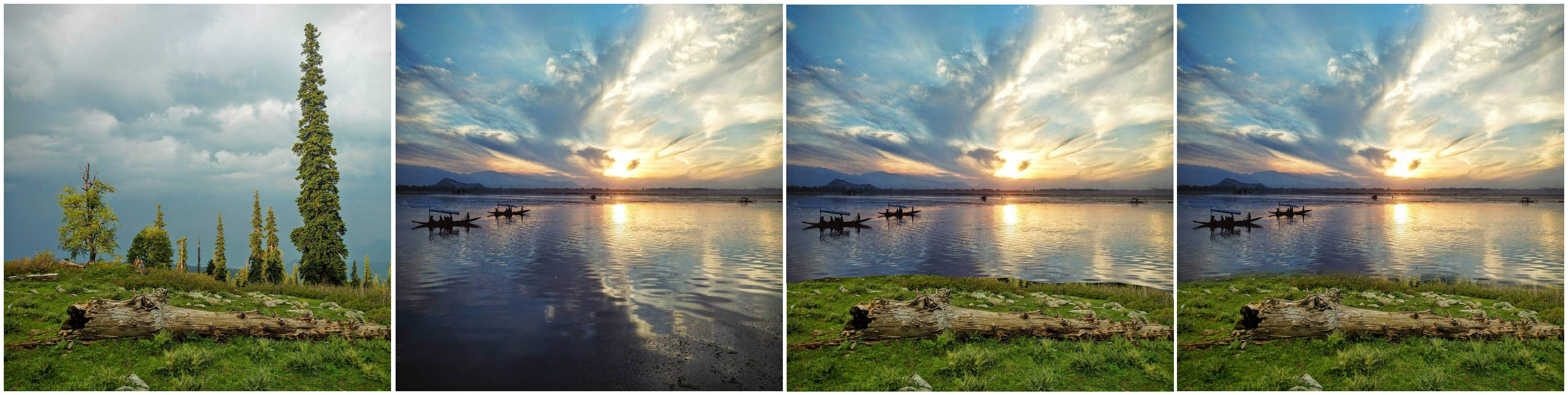}
       
        \caption{First and second column: input image; Third column: image generated by naively copying the left half from the first image and the right half from the second image; Fourth column: image generated by our extended embedding algorithm.}
        \label{fig:latent1}
    \end{figure}   

\section{Frequently Used Building Blocks}
\label{ref:fub}
We identify four fundamental building blocks that are used in multiple applications described in Sec.~\ref{sec:spp}. While terms of the loss function can be controlled by spatial masks ($M_s, M_m, M_p$), we also use binary masks $w_m$ and $n_m$ to indicate what subset of variables should be optimized during an optimization process. For example, we might set $w_m$ to only update the $w$ variables corresponding to the first $k$ layers. In general, $w_m$ and $n_m$ contain $1$s for variables that should be updated and $0$s for variables that should remain constant. In addition to the listed parameters, all building blocks need initial variable values $w_{ini}$ and $n_{ini}$. For all experiments, we use a 32GB Nvidia V100 GPU.\\

\textbf{Masked $W^{+}$ optimization ($W_{l}$):}
This function optimizes $w \in W^{+}$, leaving $n$ constant. We use the following parameters in the loss function (L) Eq. \ref{eq:loss_function}: $\lambda_s = 0$, $\lambda_{mse_{1}} = 10^{-5}$, $\lambda_{mse_{2}} = 0$, $\lambda_{p} = 10^{-5}$.
We denote the function as:
\begin{equation}
\begin{multlined}
W_{l}(M_{p},M_m,w_{m},w_{ini},n_{ini},x) = \\ \argmin_{w_{m}} \lambda_p L_{percept}(M_{p}, G(w, n), x) + \\ \frac{\lambda_{mse_{1}}}{N} \|M_{m} \odot (G(w, n) - x)\|_2^2
\end{multlined}
\label{eq:loss_function1}
\end{equation}
where $w_{m}$ is a mask for $W^{+}$ space. We either use Adam~\cite{kingma2014adam} with learning rate 0.01 or gradient descent with learning rate 0.8, depending on the application. Some common settings for Adam are: $\beta_1=0.9$, $\beta_2=0.999$, and $\epsilon=1e^{-8}$. In Sec.~\ref{sec:spp}, we use Adam unless specified.\\  

\textbf{Masked Noise Optimization (${Mk}_{n}$):}
This function optimizes $n \in N_s$, leaving $w$ constant. The Noise space $N_s$ has dimensions $\left\{\mathbb{R}^{4 \times 4}, \ldots, \mathbb{R}^{1024 \times 1024}\right\}$. In total there are 18 noise maps, two for each resolution.
We set following parameters in the loss function (L) Eq. \ref{eq:loss_function}: $\lambda_s = 0$, $\lambda_{mse_{1}} =10^{-5}$, $\lambda_{mse_{2}} = 10^{-5}$, $\lambda_{p} = 0$.
We denote the function as:
     \begin{equation} 
    \begin{multlined}
     {Mk}_{n}(M,w_{ini},n_{ini},x,y) = \\ \argmin_{n} \frac{\lambda_{mse_{2}}}{N} \|M_{m} \odot (G(w,n) - x)\|_2^2 +\\ \frac{ \lambda_{mse_{1}}}{N} \|(1 - M_{m}) \odot (G(w,n) - y)\|_2^2
     \end{multlined}
    \label{eq:loss_func}
\end{equation}
For this optimization, we use Adam with learning rate 5, $\beta_1=0.9$, $\beta_2=0.999$, and $\epsilon=1e^{-8}$. Note that the learning rate is very high.\\

\textbf{Masked Style Transfer($M_{st}$):}
 This function optimizes $w$ to achieve a given target style defined by style image $y$. We set following parameters in the loss function (L) Eq. \ref{eq:loss_function}:
$\lambda_s = 5\times10^{-7}$, $\lambda_{mse_{1}} = 0$, $\lambda_{mse_{2}} = 0$, $\lambda_{p} = 0$.
We denote the function as:    
  \begin{equation} 
    \begin{multlined}
   M_{st}(M_{s},w_{ini},n_{ini},y) = \\  \argmin_{w} \lambda_s L_{style} (M_s, G(w,n), y)
     \end{multlined}
    \label{eq:loss_function2}
\end{equation}
    where $w$ is the whole $W^{+}$ space. For this optimization, we use Adam with learning rate 0.01, $\beta_1=0.9$, $\beta_2=0.999$, and $\epsilon=1e^{-8}$. \\
   
\textbf{Masked activation tensor operation ($I_{att}$):}
    This function describes an activation tensor operation. Here, we represent the generator $G(w,n,t)$ as a function of $W^{+}$ space variable $w$, Noise space variable $n$, and input tensor $t$. The operation is represented by: 
     \begin{equation} 
     \begin{multlined}
     I_{att}(M_{1},M_{2},w,n_{ini},l) = \\ G(w,n , M_{1} \odot (A_{l}^{I_{1}}) + (1-M_{2}) \odot  (B_{l}^{I_{2}}))
       \end{multlined}
    \label{eq:loss_function5}
    \end{equation}
        where $A_{l}^{I_{1}}$ and $B_{l}^{I_{2}}$ are the activations corresponding to images $I_{1}$ and $I_{2}$ at layer $l$, and $M_{1}$ and $M_{2}$ are the masks downsampled using nearest neighbour interpolation to match the $H_{l}\times W_{l}$ resolution of the activation tensors. 

\section{Applications}
\label{sec:spp}

In the following we describe various applications enabled by our framework.

\begin{algorithm}[h]
\SetAlgoLined{
 \KwIn{image $I_m \in \mathbb{R}^{n \times m \times 3}$}
 \KwOut{the embedded code $(w_{out},n_{out})$}
$(w_{ini},n_{ini}) \leftarrow$ initialize()\; 
$w_{out} = W_{l}(1,1,1,w_{ini},n_{ini},I_m)$\;
$n_{out} = {Mk}_{n}(1,w_{out},n_{ini},I_{m},0)$\;
 }
 \caption{Improved Image Reconstruction}
 \label{alg:latent_space_embedding}
\end{algorithm}

\subsection{Improved Image Reconstruction}
\label{sec:i}
As shown in Fig.~\ref{fig:Latent_noise}, any image can be embedded by optimizing for variables $w \in W^{+}$ and $n \in N_s$. Here we describe the details of this embedding (See Alg.~\ref{alg:latent_space_embedding}). First, we initialize:
$w_{ini}$ is a mean face latent code \cite{STYLEGAN2018} or random code sampled from $U[-1,1]$ depending on whether the embedding image is a face or a non-face, and $n_{ini}$ is sampled from a standard normal distribution $N(0,I)$~\cite{STYLEGAN2018}. 
Second, we apply masked $W^{+}$ optimization ($W_{l}$) without using spatial masks or masking variables. That means all masks are set to $1$. $I_m$ is the target image we try to reconstruct. Third, we perform masked noise optimization (${Mk}_{n}$), again without making use of masks.
The images reconstructed are of high fidelity. The PNSR score range of 39 to 45 dB provides an insight of how expressive the Noise space in StyleGAN is. Unlike the $W^{+}$ space, the Noise space is used for spatial reconstruction of high frequency features. We use 5000 iterations of $W_{l}$ and 3000 iterations of ${Mk}_{n}$ to get PSNR scores of 44 to 45 dB. Additional iterations did not improve the results in our tests.

        \begin{figure}
        \centering
        \includegraphics[width=\linewidth]{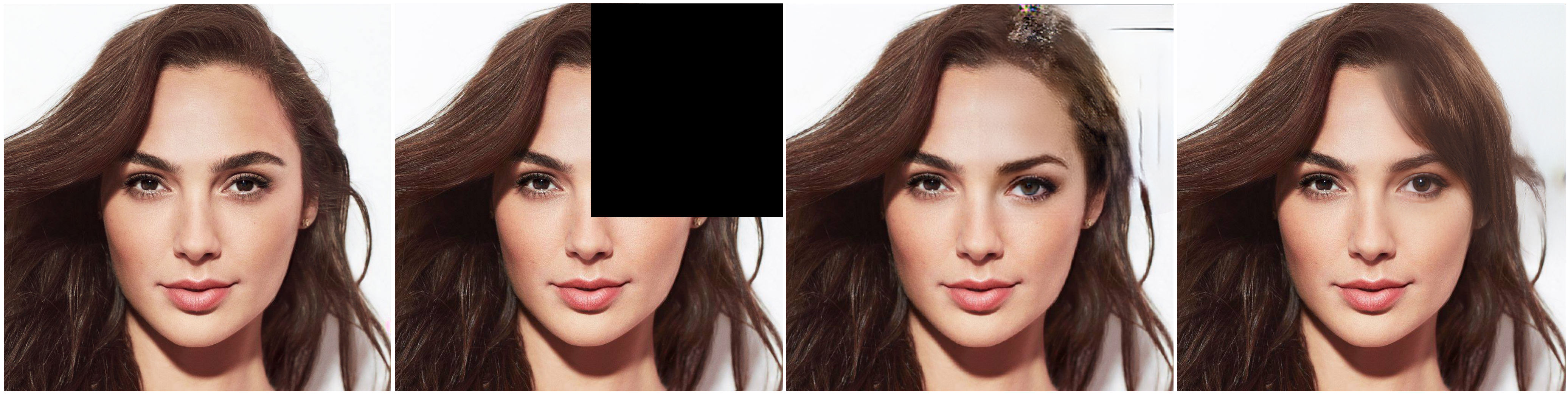}
        \includegraphics[width=\linewidth]{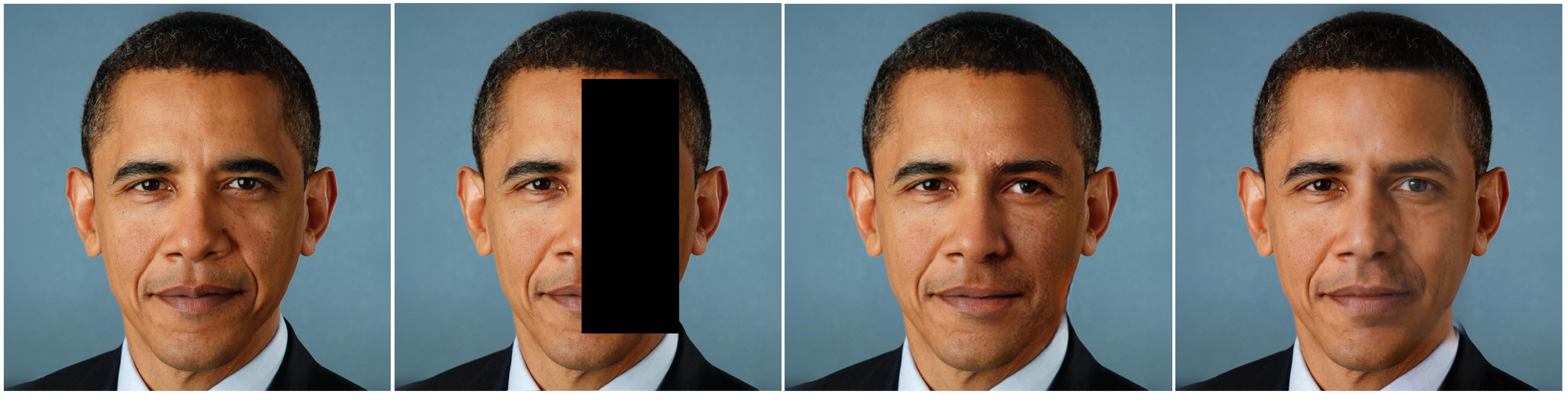}
        \includegraphics[width=\linewidth]{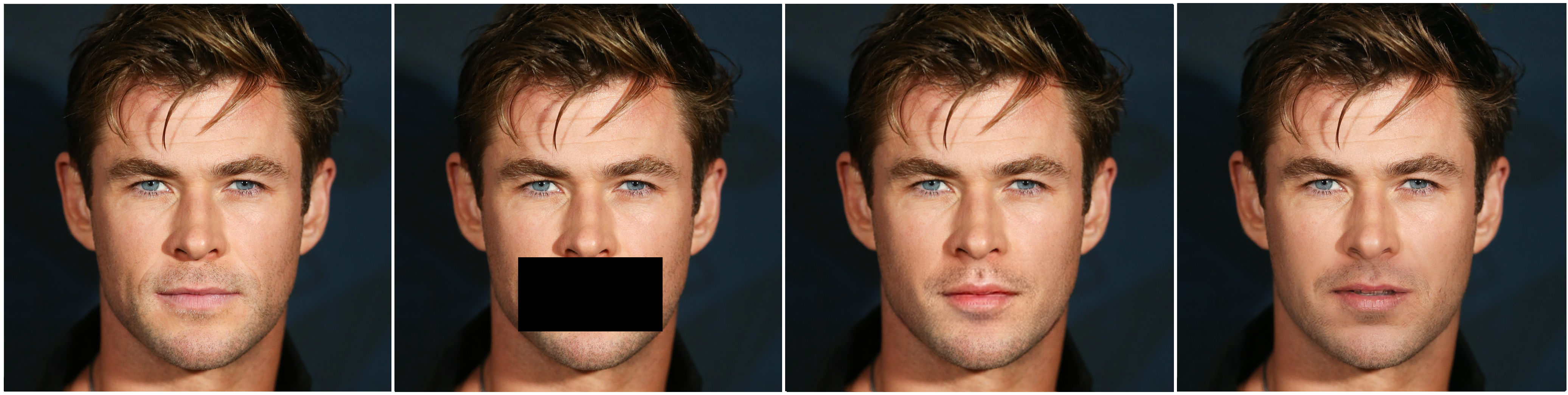}
        \caption{First column: original image; Second column: defective image ; Third column: inpainted image via partial convolutions~\cite{Liu_2018}; Fourth column: inpainted image using our method. }
        \label{fig:imp1}
    \end{figure}
     \begin{figure}
        \centering
        \includegraphics[width=\linewidth]{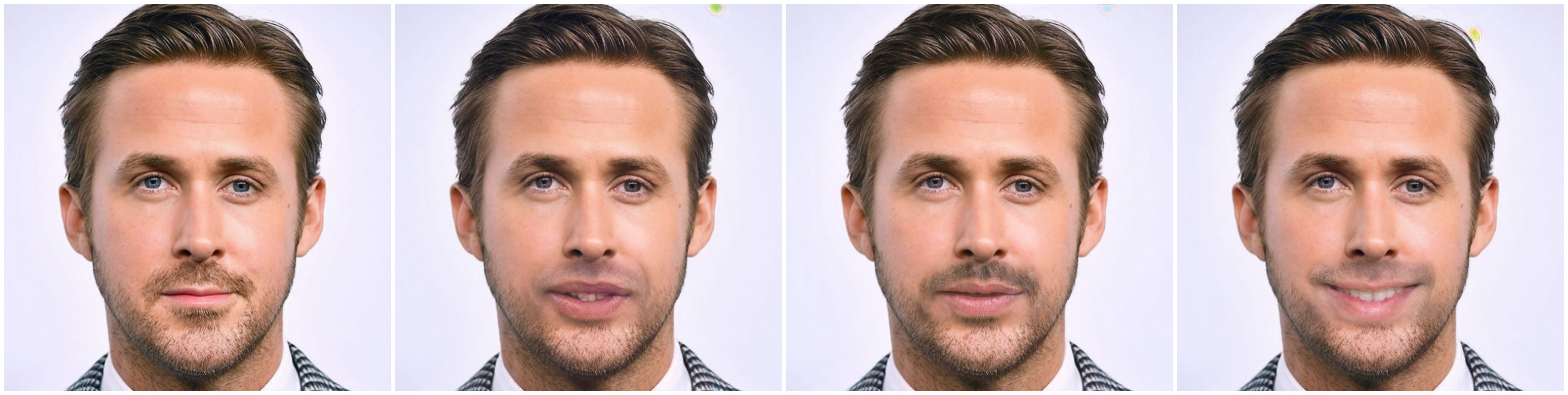}
        \includegraphics[width=\linewidth]{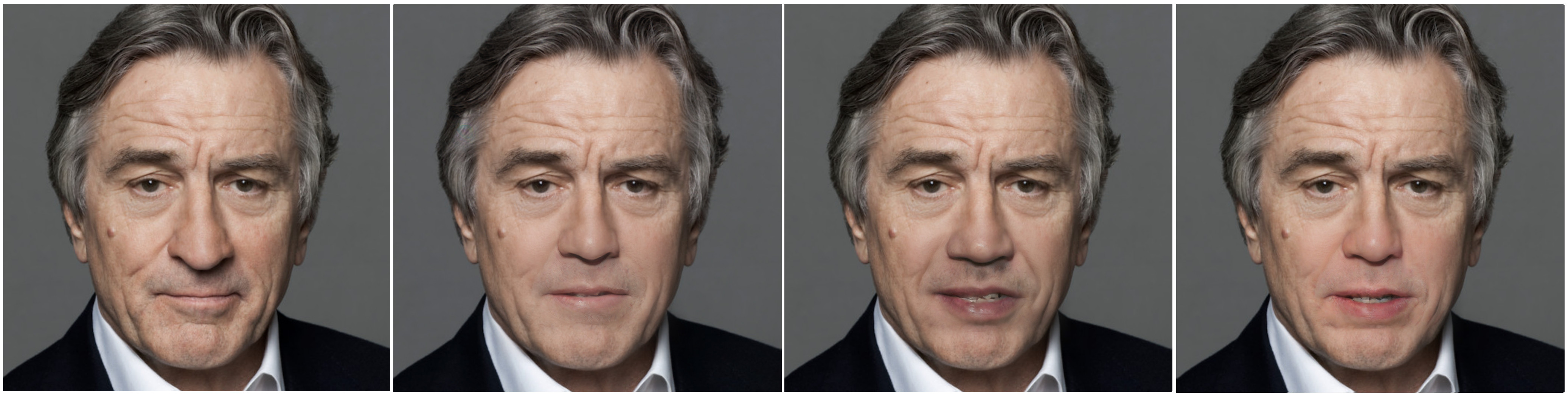}
        \caption{Inpainting using different initializations $w_{ini}$.}
        \label{fig:imp2}
    \end{figure}

  \begin{figure*}[t]
       \vspace{-0.5cm} 
        \centering
        \begin{subfigure}{0.49\textwidth}
            \includegraphics[width= 0.99\linewidth]{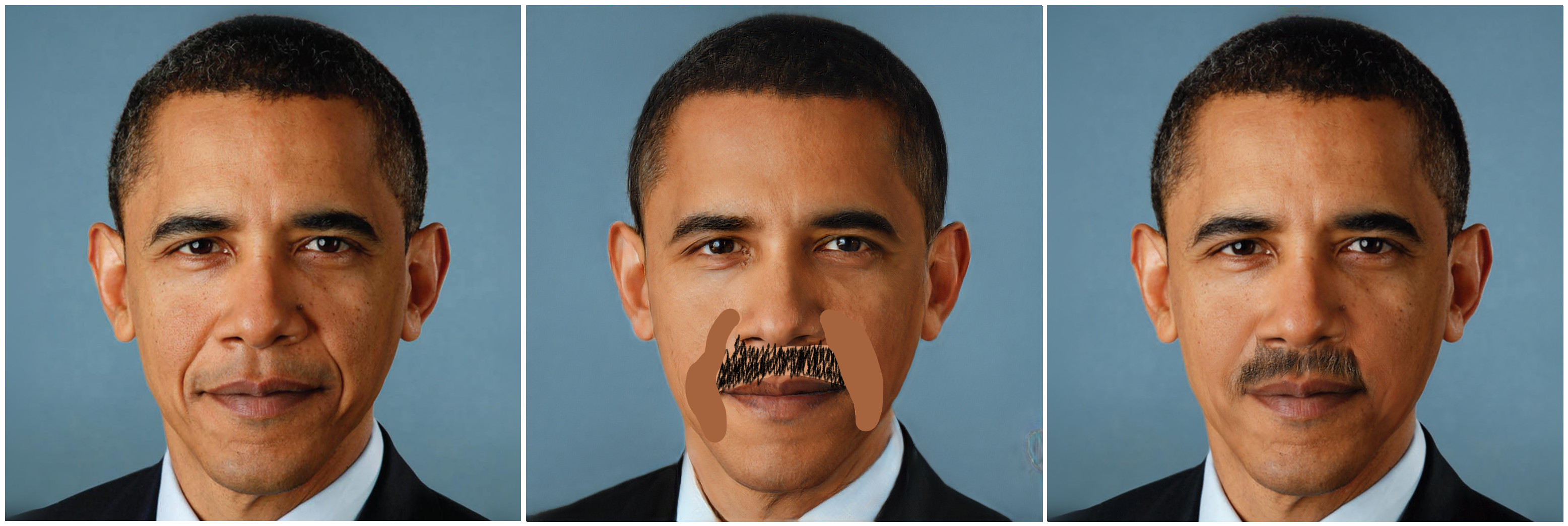}
             \includegraphics[width=0.99\linewidth]{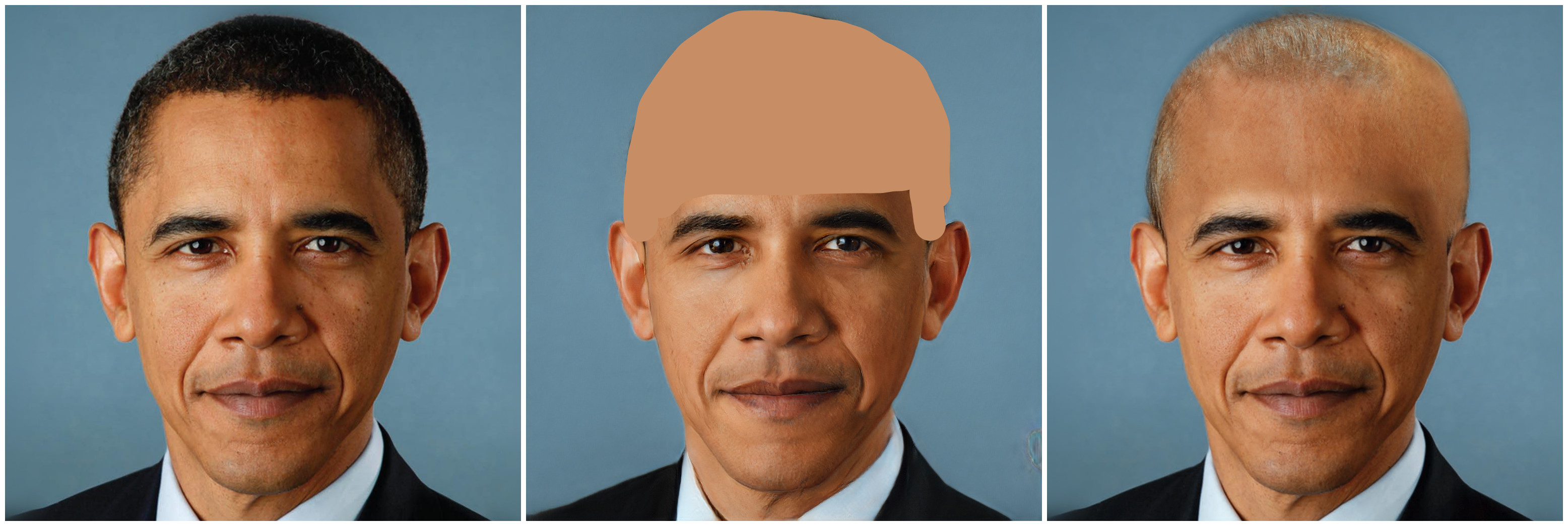}
             \includegraphics[width=0.99\linewidth]{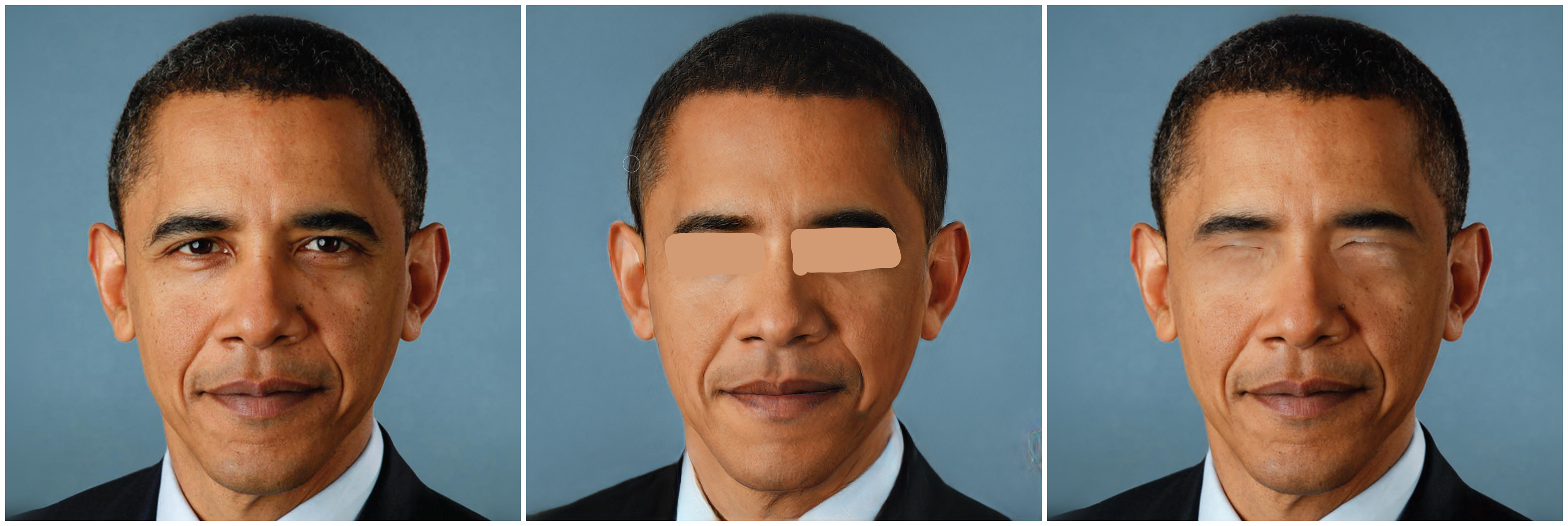}
             
        \end{subfigure}
        \begin{subfigure}{0.49\textwidth}
             \includegraphics[width=0.99\linewidth]{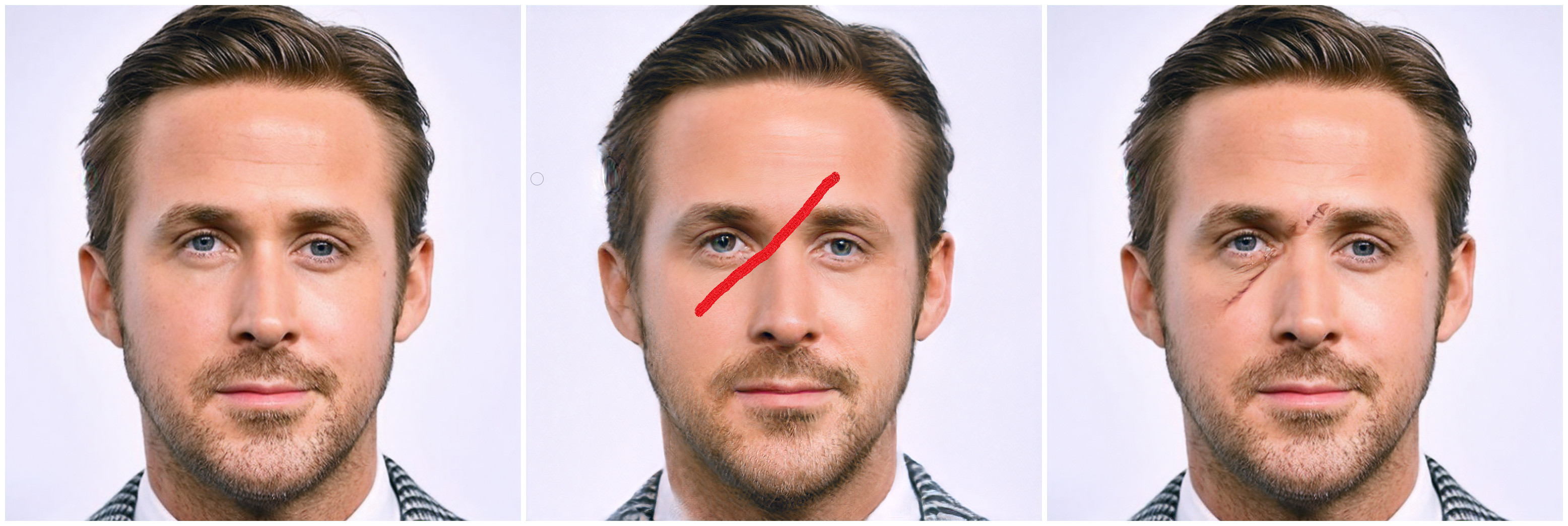}
              \includegraphics[width=0.99\linewidth]{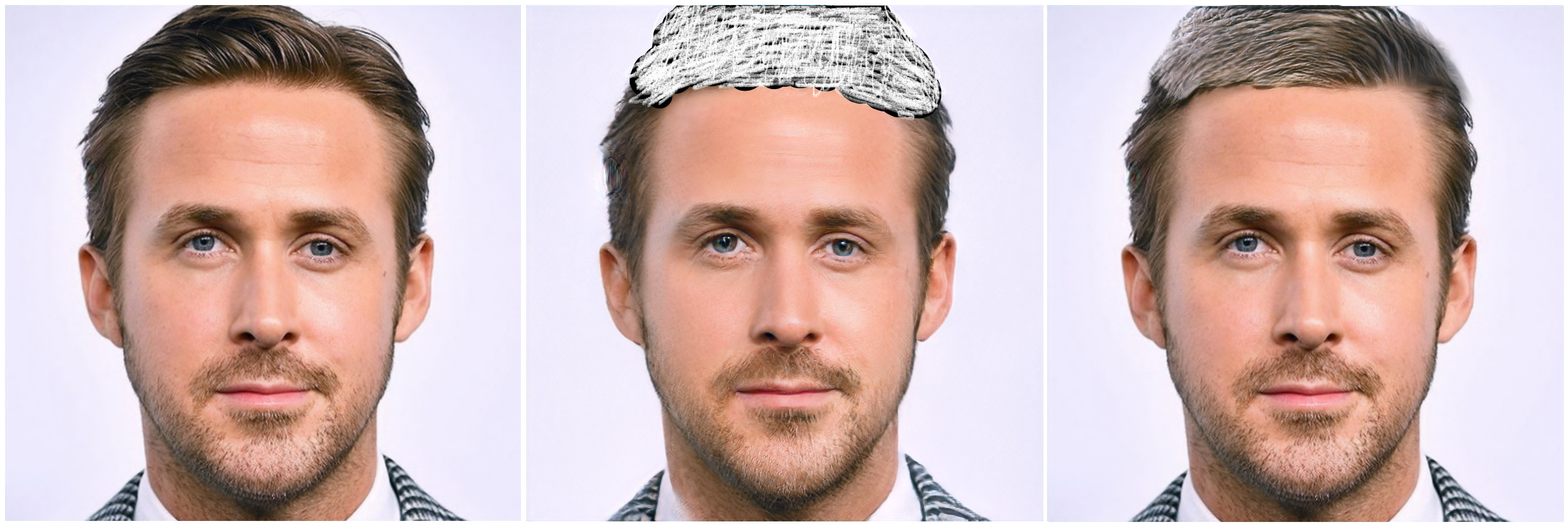}
               \includegraphics[width=0.99\linewidth]{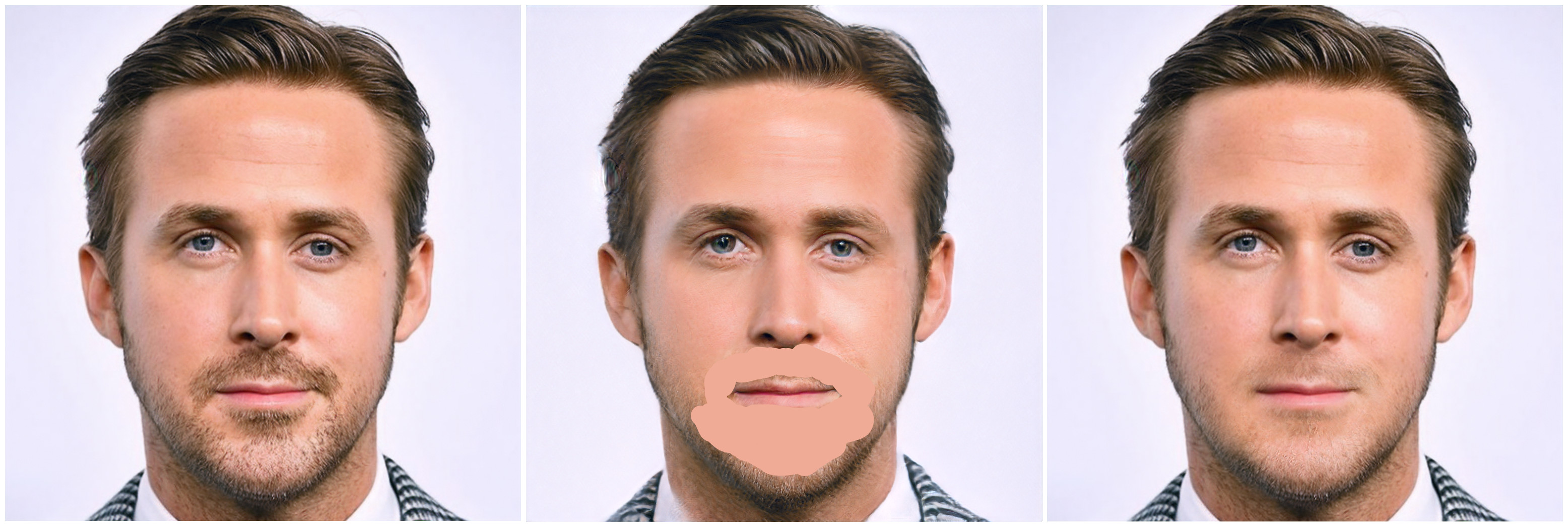}
        \end{subfigure}
        \caption{Column 1 \& 4: base image; Column 2 \& 5: scribbled image ; Column 3 \& 6: result of local edits.}
        \label{fig:latent4}
    \end{figure*}

\subsection{Image Crossover}
\label{sec:cross}
\begin{algorithm}[h]
\SetAlgoLined
 \KwIn{images $I_1, I_2 \in \mathbb{R}^{n \times m \times 3}$; mask $M_{blur}$}
 \KwOut{the embedded code $(w_{out},n_{out})$}
 $(w^*,n_{ini}) \leftarrow$ initialize()\;
{$w_{out} = W_{l}(M_{blur},M_{blur},1,w^*,n_{ini},I_1)$\ $+ W_{l}(1-M_{blur},1-M_{blur},1,w^*,n_{ini},I_2)$\;
$n_{out} = {Mk}_{n}(M_{blur},w_{out},n_{ini},I_1,I_2)$\;
 }
 \caption{Image Crossover}
 \label{alg:image_crossover1}
\end{algorithm}
We define the image crossover operation as copying parts from a source image $y$ into a target image $x$ and blending the boundaries. 
As initialization, we embed the target image $x$ to obtain the $W^{+}$ code $w^{*}$. We then perform masked $W^{+}$ optimization ($W_{l}$) with blurred masks $M_{blur}$ to embed the regions in $x$ and $y$ that contribute to the final image. Blurred masks are obtained by convolution of the binary mask with a Gaussian filter of suitable size. Then, we perform noise optimization. Details are provided in Alg.~\ref{alg:image_crossover1}.

Other notations are the same as described in Sec~\ref{sec:i}. Fig.~\ref{fig:latent1} and Fig.~\ref{fig:teaser} show example results. 
We deduce that the reconstruction quality of the images is quite high. 
For the experiments, we use 1000 iterations in the function masked $W^{+}$ optimization and 1000 iterations in ${Mk}_{n}$.

\subsection{Image Inpainting}

\begin{algorithm}[h]
\SetAlgoLined
 \KwIn{image $I_{def} \in \mathbb{R}^{n \times m \times 3}$; masks $M,M_{blur+}$}
 \KwOut{the embedded code $(w_{out},n_{out})$}
 $(w_{ini},n_{ini}) \leftarrow$ initialize()\;

{$w_{out} = W_{l}(1-M,1-M,w_m,w_{ini},n_{ini},I_{def})$\;
$n_{out} = {Mk}_{n}(1-M_{blur+},w_{out},n_{ini},I_{def},G(w_{out}))$\;
 }
 \caption{Image Inpainting}
 \label{alg:latent_space_embedding2}
\end{algorithm}

In order to perform a semantically meaningful inpainting, we embed into the early layers of the $W^{+}$ space to predict the missing content and in the later layers to maintain color consistency. We define the image $x$ as a defective image ($I_{def}$). Also, we use the mask $w_m$ where the value is 1 corresponding to the first 9 (1 to 9), $17^{th}$ and $18^{th}$ layer of $W^+$. As an initialization, we set $w_{ini}$ to the mean face latent code~\cite{STYLEGAN2018}. We consider $M$ as the mask describing the defective region. Using these parameters, we perform the masked $W^+$ optimization $W_{l}$. Then we perform the masked noise optimization ${Mk}_{n}$ using $M_{blur+}$ which is the slightly larger blurred mask used for blending. Here $\lambda_{mse_{2}}$ is taken to be $10^{-4}$. Other notations are the same as described in Sec~\ref{sec:i}. Alg.~\ref{alg:latent_space_embedding2} shows the details of the algorithm. We perform 200 steps of gradient descent optimizer for masked $W^+$ optimization $W_l$ and 1000 iterations of masked noise optimization ${Mk}_{n}$. Fig.\ref{fig:imp1} shows example inpainting results. The results are comparable with the current state of the art, partial convolution~\cite{Liu_2018}. The partial convolution method frequently suffers from regular artifacts (see Fig.\ref{fig:imp1} (third column)). These artifacts are not present in our method. In Fig.\ref{fig:imp2} we show different inpainting solutions for the same image achieved by using different initializations of $w_{ini}$ , which is an offset to mean face latent code sampled independently from a uniform distribution $U[-0.4,0.4]$. The initialization mainly affects layers 10 to 16 that are not altered during optimization. Multiple inpainting solutions cannot be computed with existing state-of-the-art methods.

       \begin{figure}
         \centering
       \includegraphics[width=\linewidth]{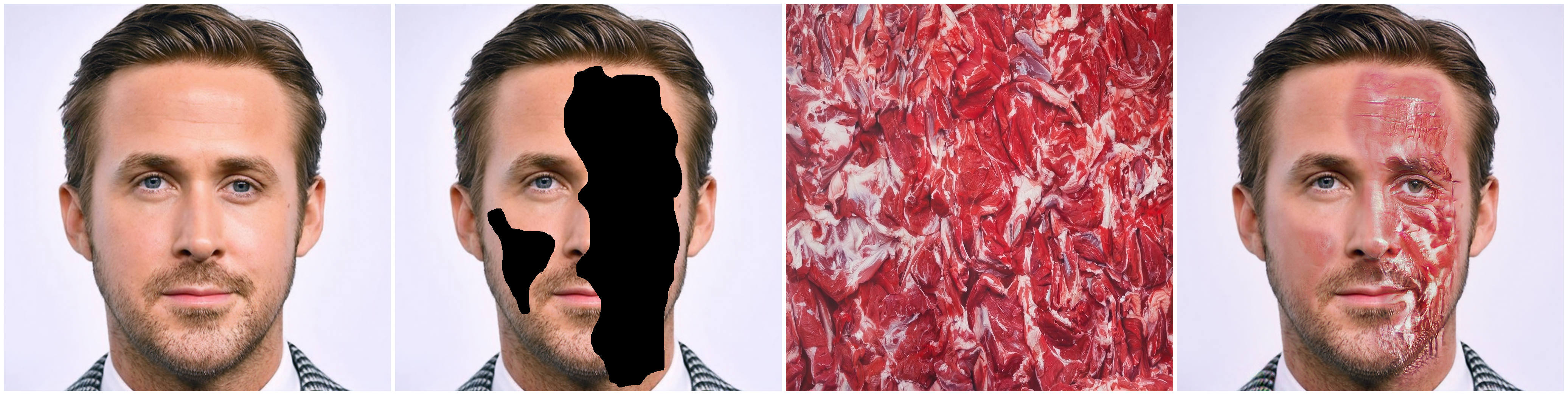}
         \includegraphics[width=\linewidth]{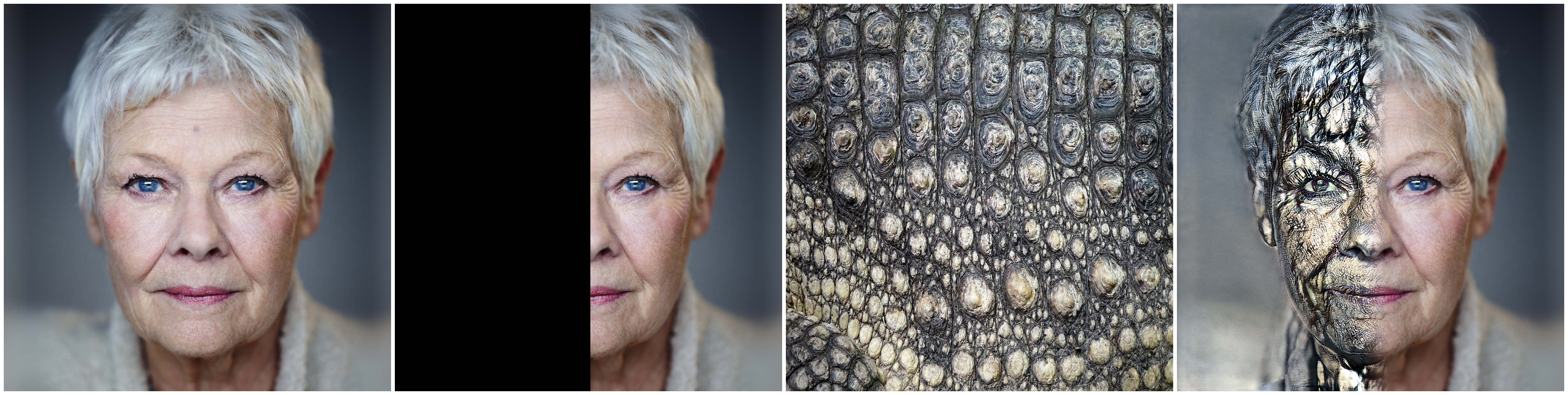}
         \caption{First column: base image; Second column: mask area; Third column: style image; Fourth column: local style transfer result. }
         \label{fig:latent_next}
    \end{figure}
    
     \begin{figure}[t]
        \centering
         \includegraphics[width=\linewidth]{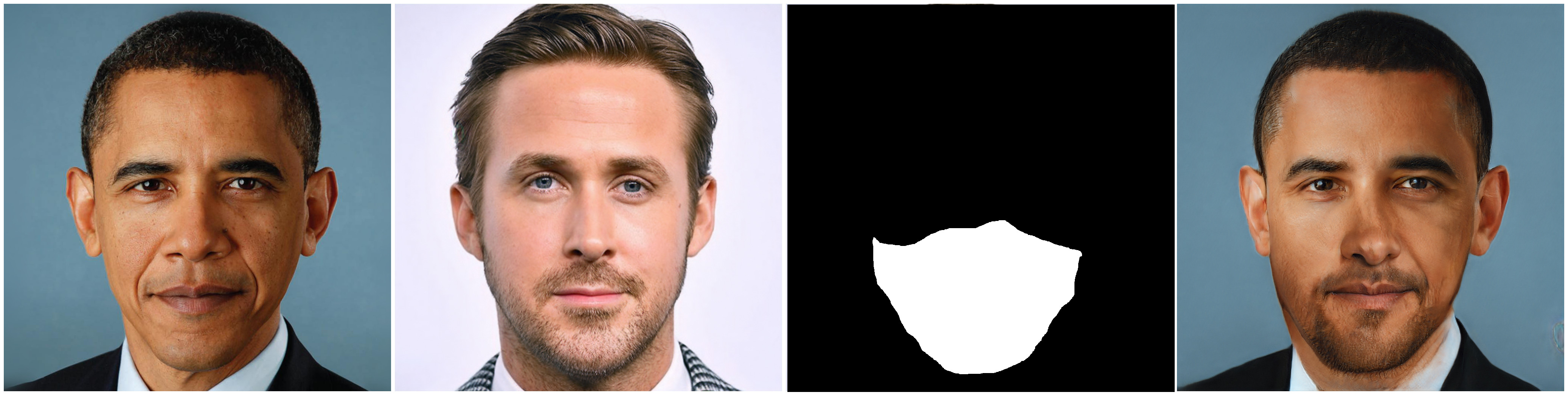}
        \includegraphics[width=\linewidth]{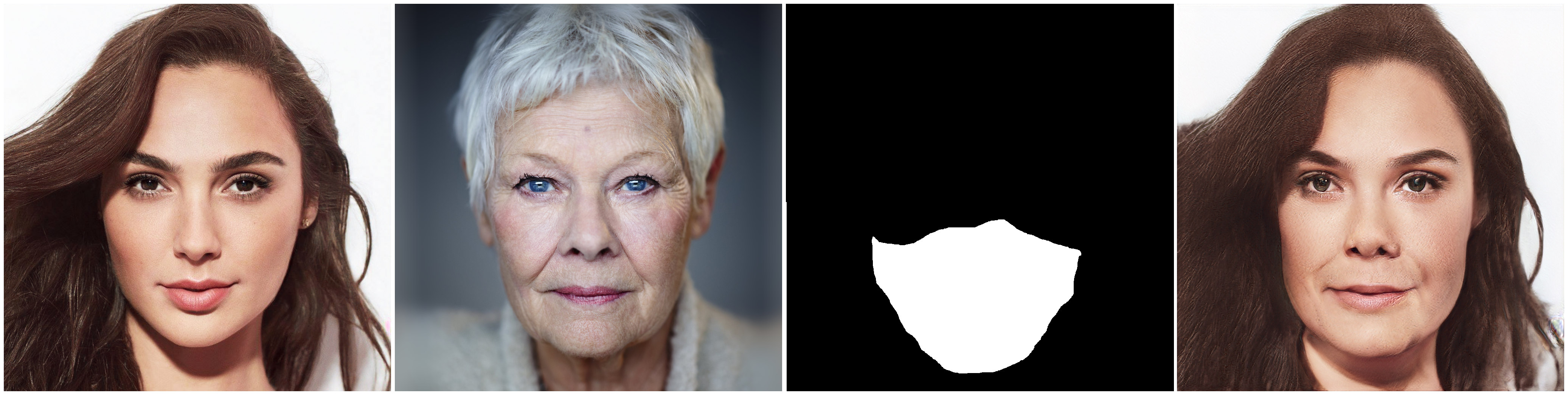}
       
        \includegraphics[width=\linewidth]{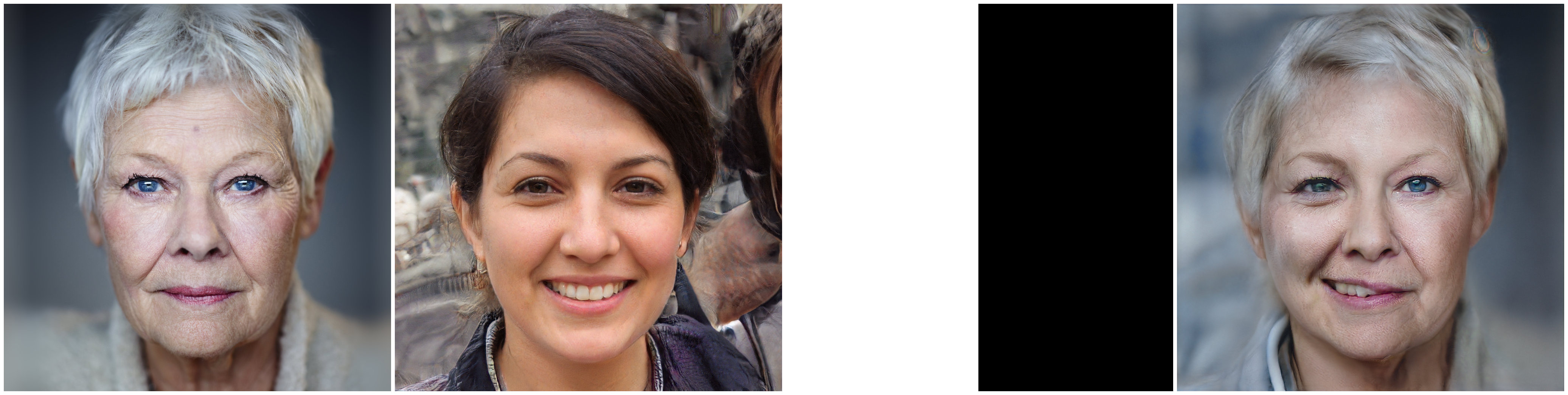}
        \includegraphics[width=\linewidth]{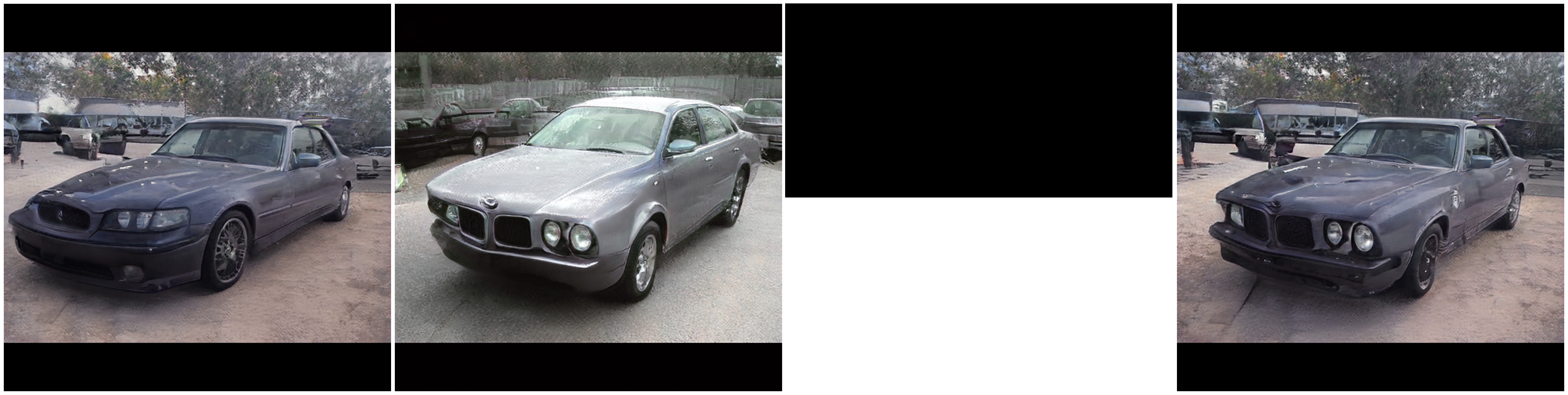}
        \includegraphics[width=\linewidth]{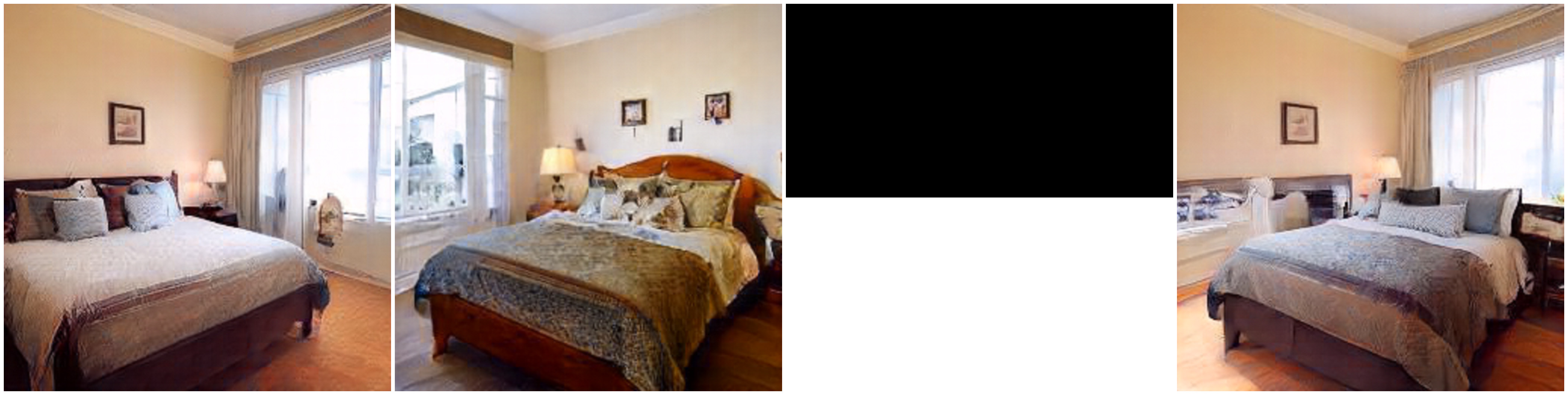}
        \caption{First column: base image; Second column: attribute image; Third column: mask area; Fourth column: image generated via attribute level feature transfer. }
        \label{fig:latent2}
    \end{figure}

\subsection{Local Edits using Scribbles}

\begin{algorithm}[h]
\SetAlgoLined
 \KwIn{image $I_{scr} \in \mathbb{R}^{n \times m \times 3}$; masks $M_{blur}$}
 \KwOut{the embedded code $(w_{out},n_{out})$}

  $(w^*,n_{ini}) \leftarrow$ initialize()\;
{$w_{out} = W_{l}(1,1,w_m,w^*,n_{ini},I_{scr})$\
$ +\lambda \|w^{*} - w_{out}\|_2$\;
$n_{out} = {Mk}_{n}(M_{blur},w_{out},n_{ini},I_{scr},G(w_{out}))$\;
 }
 \caption{Local Edits using Scribble}
 \label{alg:latent_space_embedding3}
\end{algorithm}
\begin{algorithm}[h]
\SetAlgoLined
 \KwIn{images $I_{1}, I_{2} \in \mathbb{R}^{n \times m \times 3}$; masks $M_{blur}$}
 \KwOut{the embedded code $(w_{out},n_{out})$}

   $(w^*,n_{ini}) \leftarrow$ initialize()\;
{$w_{out} = W_{l}(M_{blur},M_{blur},1 ,w^*,n_{ini},I_{1})$\
$+ M_{st}(1-M_{blur},w^{*} ,n_{ini}, I_{2})$\;
$n_{out} = {Mk}_{n}(M_{blur},w_{out},n_{ini},I_{1},G(w_{out}))$\;
 }
 \caption{Local Style Transfer}
 \label{alg:latent_space_embedding4}
\end{algorithm}
Another application is performing semantic local edits guided by user scribbles. We show that simple scribbles can be converted to photo-realistic edits by embedding into the first 4 to 6 layers of $W^+$ (See Fig.\ref{fig:latent4}). This enables us to do local edits without training a network. We define an image $x$ as a scribble image ($I_{scr}$). Here, we also use the mask $w_m$ where the value is 1 corresponding to the first 4,5 or 6 layers of the $W^+$ space. As initialization, we set the $w_{ini}$ to $w^{*}$ which is the $W^+$ code of the image without scribble. We perform masked $W^+$ optimization using these parameters. Then we perform masked noise optimization ${Mk}_{n}$ using $M_{blur}$. Other notations are the same as described in Sec~\ref{sec:i}. Alg.~\ref{alg:latent_space_embedding3} shows the details of the algorithm. We perform 1000 iterations using Adam with a learning rate of 0.1 of masked $W^{+}$ optimization $W_{l}$ and then 1000 steps of masked noise optimization ${Mk}_{n}$ to output the final image.
\subsection{Local Style Transfer}

Local style transfer modifies a region in the input image $x$ to transform it to the style defined by a style reference image. First, we embed the image in $W^+$ space to obtain the code $w^*$. Then we apply the masked $W^+$ optimization $W_l$ along with masked style transfer $M_{st}$ using blurred mask $M_{blur}$. Finally, we perform the masked noise optimization ${Mk}_{n}$ to output the final image. Alg.~\ref{alg:latent_space_embedding4} shows the details of the algorithm. Results for the application are shown in Fig.\ref{fig:latent_next}. We perform 1000 steps to obtain of $W_l$ along with $M_{st}$ and then perform 1000 iterations of ${Mk}_{n}$.

\subsection{Attribute level feature transfer}
\label{sec:alft}
 We extend our work to another application using tensor operations on the images embedded in $W^{+}$ space. In this application we perform the tensor manipulation corresponding to the tensors at the output of the $4^{th}$ layer of StyleGAN. We feed the generator with the latent codes ($w$, $n$) of two images $I_{1}$ and $I_{2}$ and store the output of the fourth layer as intermediate activation tensors $A_{l}^{I_{1}}$ and $B_{l}^{I_{2}}$.
A mask $M_{s}$ specifies which values to copy from $A_{l}^{I_{1}}$ and which to copy from $B_{l}^{I_{2}}$. The operation can be denoted by $I_{att}(M_{s},M_{s},w,n_{ini},4)$. In Fig.\ref{fig:latent2}, we show results of the operation. A design parameter of this application is what style code to use for the remaining layers. In the shown example, the first image is chosen to provide the style. Notice, in column 2 of Fig.\ref{fig:latent2}, in-spite of the different alignment of the two faces and objects, the images are blended well. We also show results of blending for the LSUN-car and LSUN-bedroom datasets. Hence, unlike global edits like image morphing, style transfer, and expression transfer \cite{abdal2019image2stylegan}, here different parts of the image can be edited independently and the edits are localized. Moreover, along with other edits, we show a video in the supplementary material that further shows that other semantic edits e.g. masked image morphing can be performed on such images by linear interpolation of $W^{+}$ code of one image at a time.

\section{Conclusion}

We proposed Image2StyleGAN++, a powerful image editing framework built on the recent Image2StyleGAN.
Our framework is motivated by three key insights: first, high frequency image features are captured by the additive noise maps used in StyleGAN, which helps to improve the quality of reconstructed images;
second, local edits are enabled by including masks in the embedding algorithm, which greatly increases the capability of the proposed framework;
third, a variety of applications can be created by combining embedding with activation tensor manipulation. 
From the high quality results presented in this paper, it can be concluded that our Image2StyleGAN++ is a promising framework for general image editing.
For future work, in addition to static images, we aim to extend our framework to process and edit videos.

\textbf{Acknowledgement} This work was supported by the KAUST Office of Sponsored Research (OSR) under Award No. OSR-CRG2018-3730.

{\small
\bibliographystyle{ieee}
\bibliography{egbib}
}
\clearpage
\section{Additional Results}
\subsection{Image Inpainting}
To evaluate the results quantitatively, we use three standard metrics, SSIM, MSE loss and PSNR score to compare our method with the state-of-the-art Partial Convolution~\cite{Liu_2018} and Gated Convolution~\cite{yu2018free} methods. 

\begin{figure}[b]
        \centering
        \includegraphics[width=\linewidth]{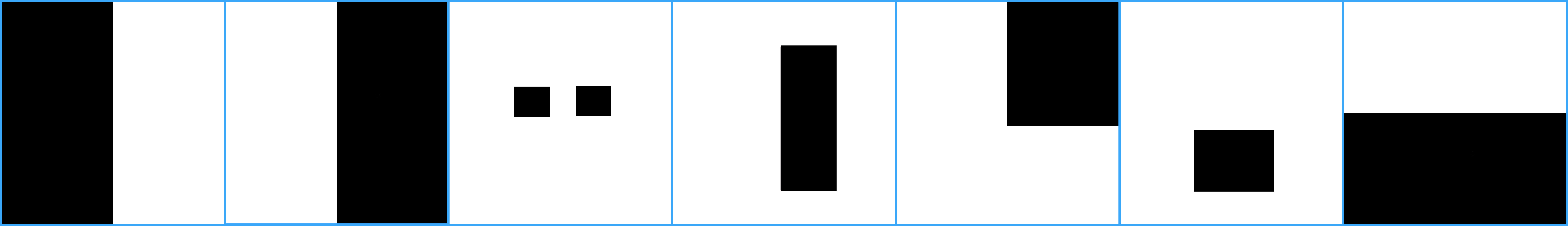}
        \caption{Masks used in the quantitative evaluation of image inpainting methods.}
        \label{fig:masks}
    \end{figure}

\begin{figure*}
        \centering
        \includegraphics[width=\linewidth]{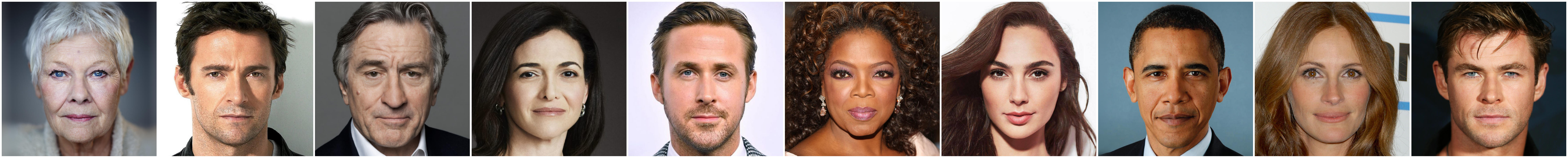}
        \caption{Images used in the quantitative evaluation of image inpainting methods.}
        
        \label{fig:all}
    \end{figure*}

As different methods produce outputs at different resolutions, we bi-linearly interpolate the output images to test the methods at three resolutions $1024\times1024$,  $512\times512$ and  $256\times256$ respectively. We use $7$ masks (Fig.~\ref{fig:masks}) and $10$ ground truth images (Fig.~\ref{fig:all}) to create $10$ defective images (\ie images with missing regions) for the evaluation. These masks and images are chosen to make the inpainting a challenging task: i) the masks are selected to contain very large missing regions, up to half of an image; ii) the ground truth images are selected to be of high variety that cover different genders, ages, races, \etc.

\begin{figure*}
     
        \centering
        \includegraphics[width=\linewidth]{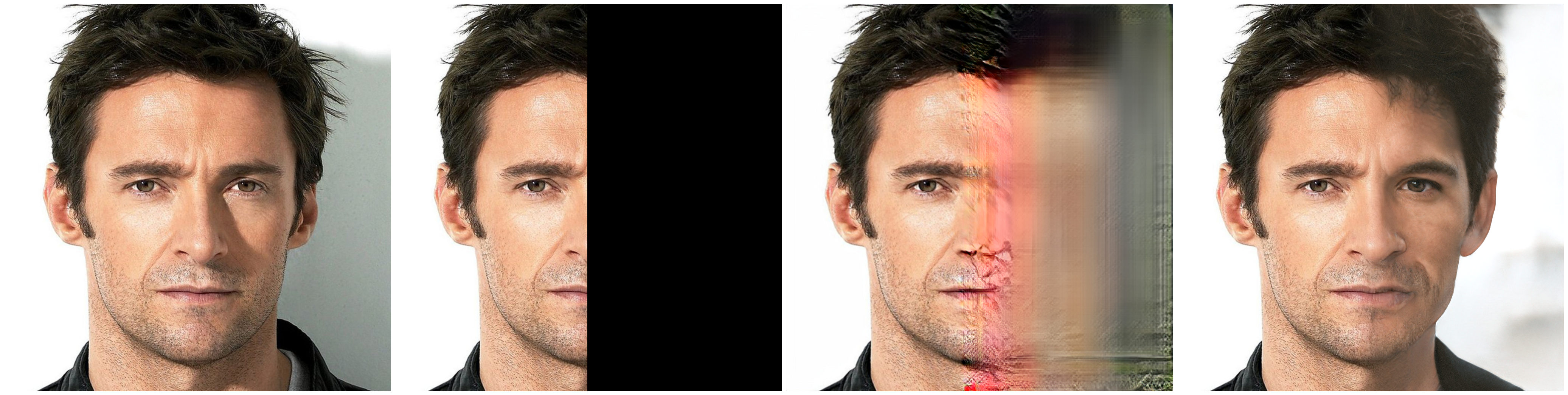}
        \includegraphics[width=\linewidth]{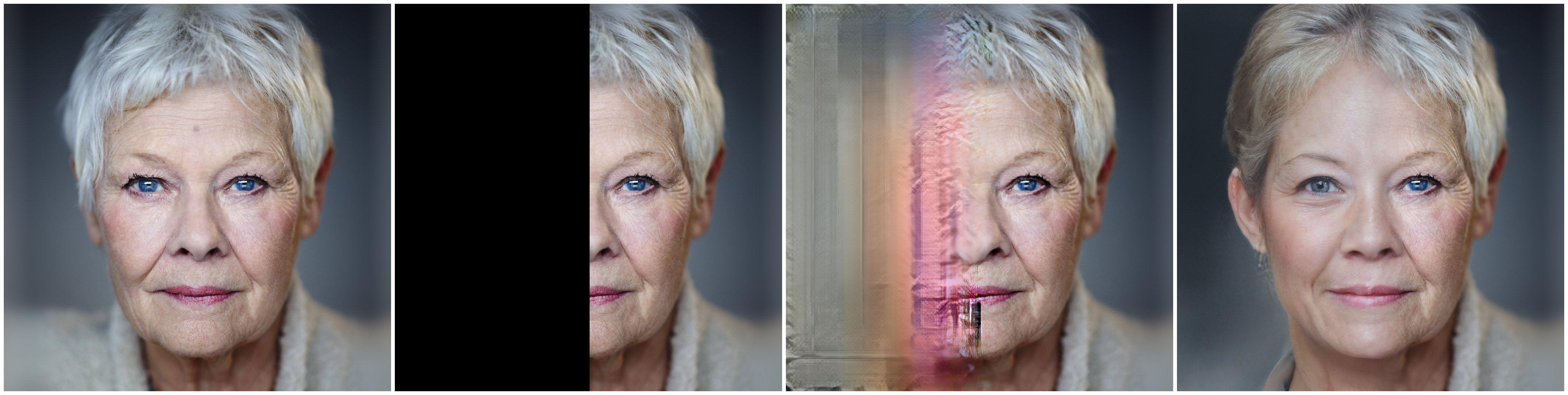}
        \includegraphics[width=\linewidth]{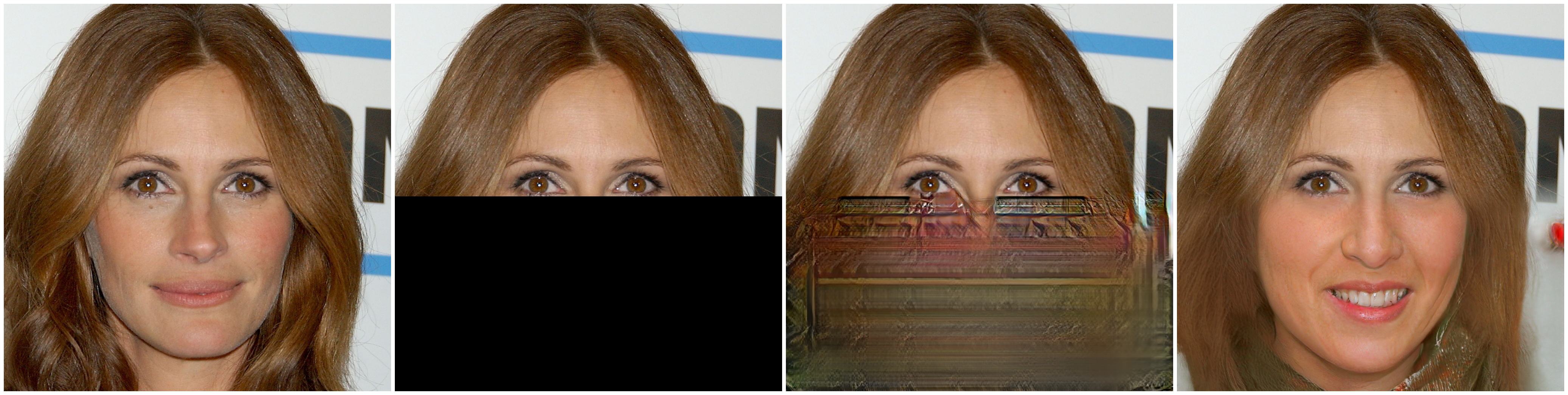}
        \includegraphics[width=\linewidth]{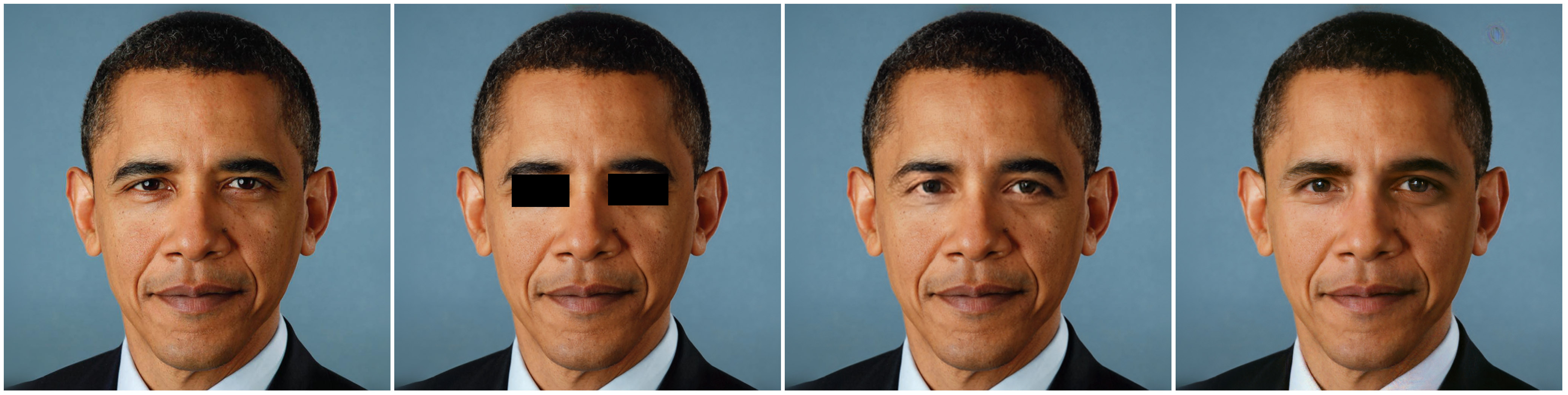}
        \includegraphics[width=\linewidth]{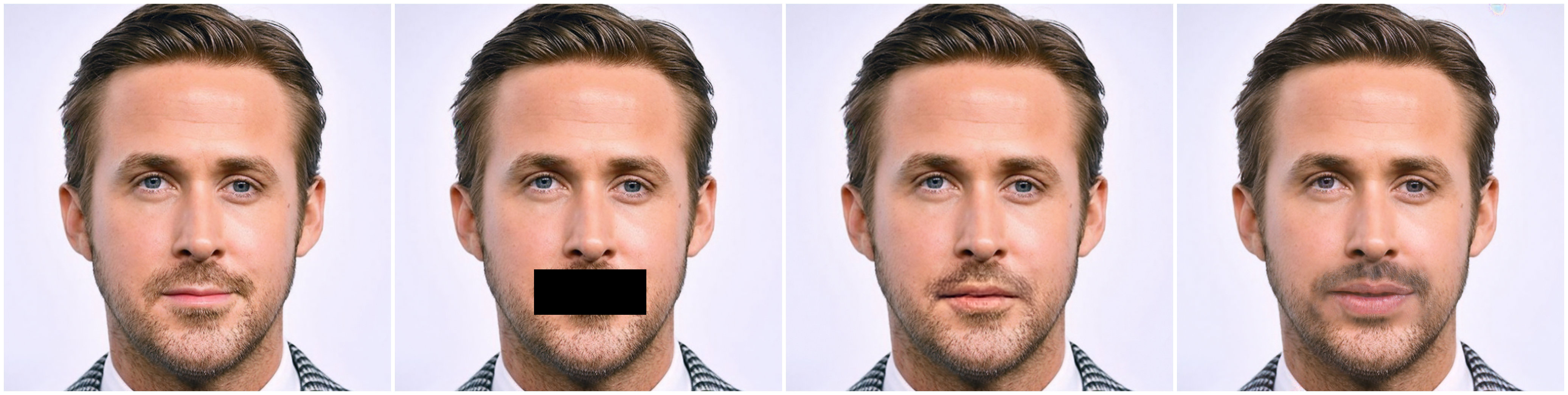}
        
        \caption{First column: original image; Second column: defective image; Third column: inpainted image via Partial Convolutions~\cite{Liu_2018}; Fourth column: inpainted image using our method.}
        \label{fig:imp_1}
    \end{figure*}

\begin{figure*}

        \centering
        \includegraphics[width=\linewidth]{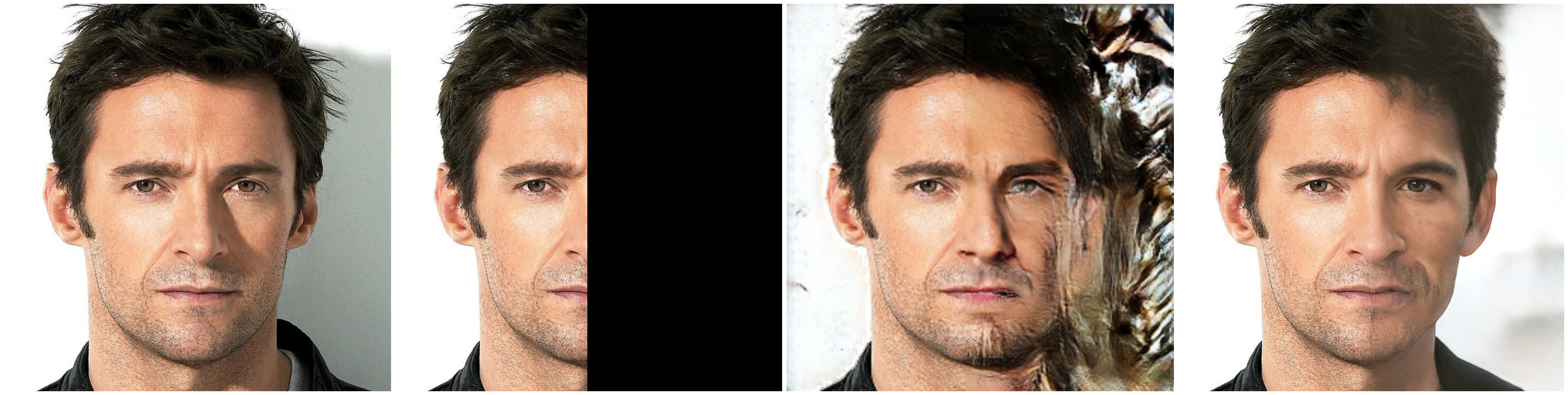}
        \includegraphics[width=\linewidth]{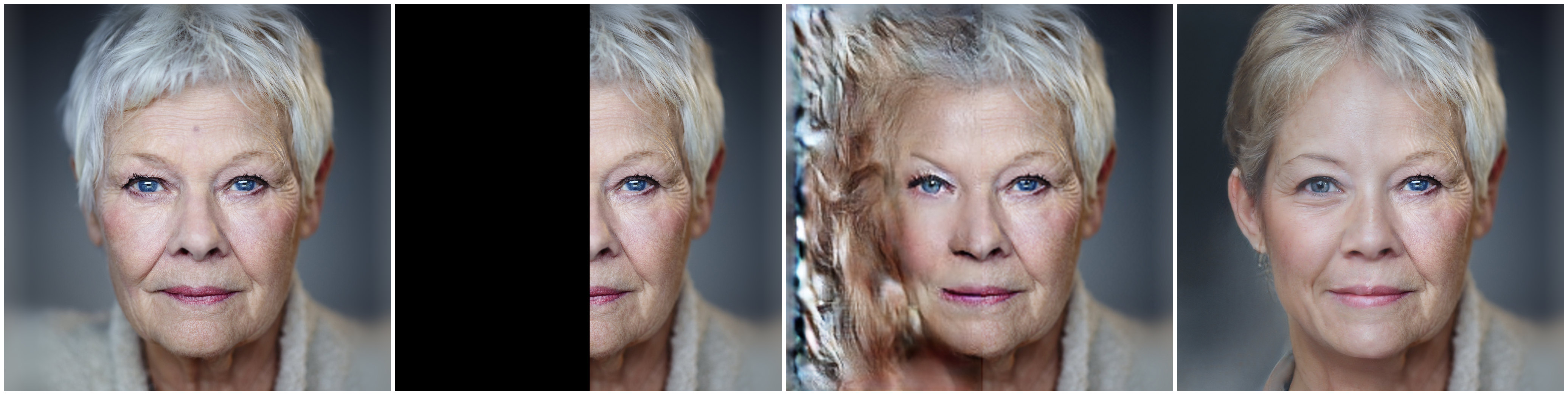}
        \includegraphics[width=\linewidth]{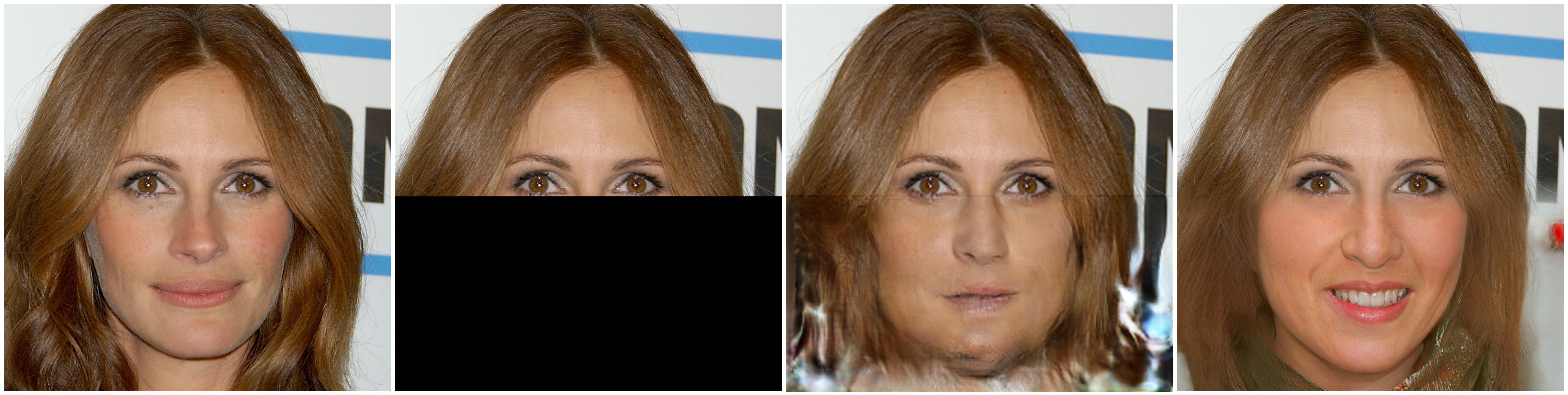}
        \includegraphics[width=\linewidth]{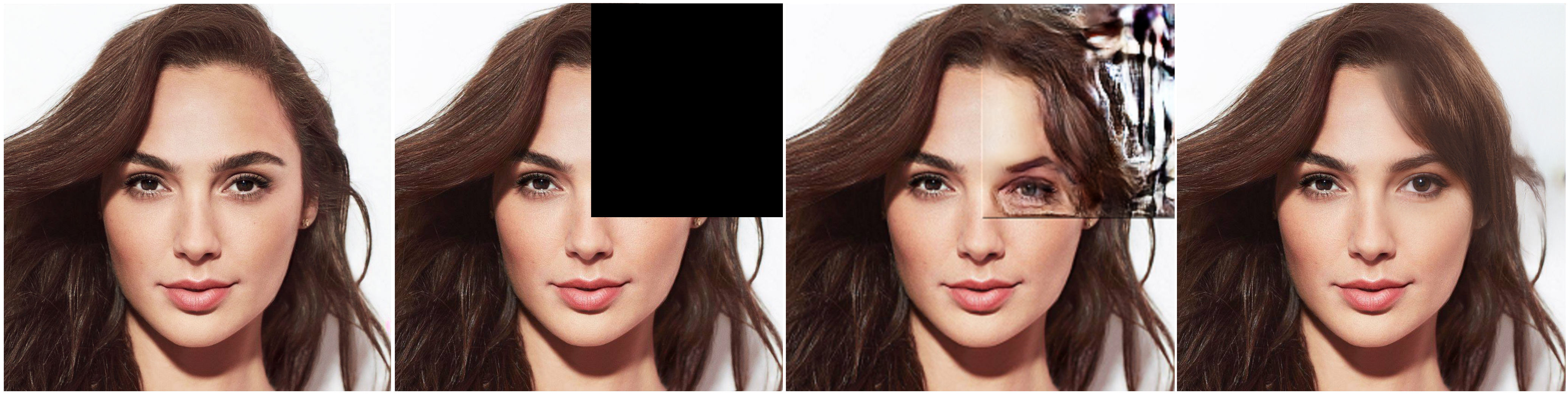}
        \includegraphics[width=\linewidth]{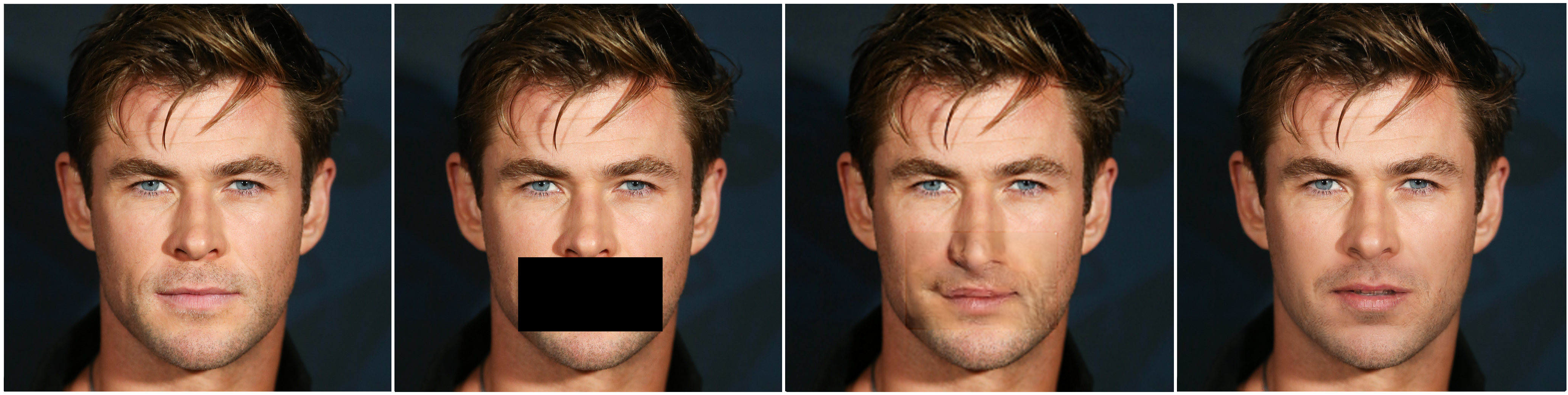}
        
        \caption{First column: original image; Second column: defective image; Third column: inpainted image via Gated Convolutions~\cite{yu2018free}; Fourth column: inpainted image using our method.}
        \label{fig:imp_2}
    \end{figure*}

Table~\ref{tab:tab1} shows the quantitative comparison results.
It can be observed that our method outperforms both Partial Convolution~\cite{Liu_2018} and Gated Convolution~\cite{yu2018free} across all the metrics. 
More importantly, the advantages of our method can be easily verified by visual inspection.
As Fig.~\ref{fig:imp_1} and Fig.~\ref{fig:imp_2} show, although previous methods (\eg Partial convolution) perform well when the missing region is small, both of them struggle when the missing region covers a significant area (\eg half) of the image. Specifically, Partial Convolution fails when the mask covers half of the input image (Fig.~\ref{fig:imp_1}); due to the relatively small resolution ($256\times256$) model, Gated Convolution can fill in the details of large missing regions, but of much lower quality compared to the proposed method (Fig.~\ref{fig:imp_2}).

\begin{figure*}[h]
        \centering
        \includegraphics[width=\linewidth]{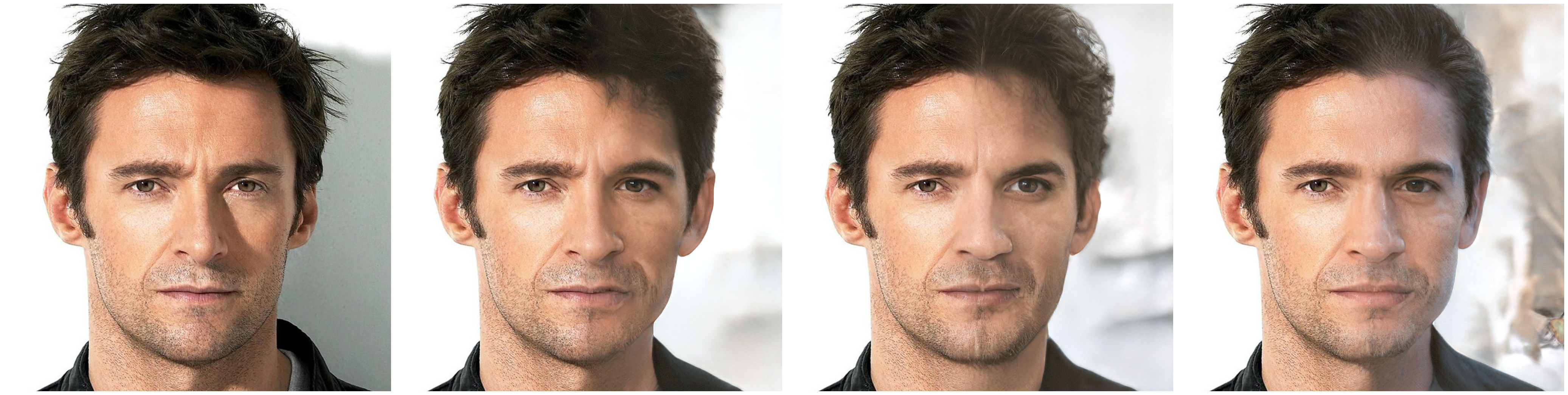}
        \includegraphics[width=\linewidth]{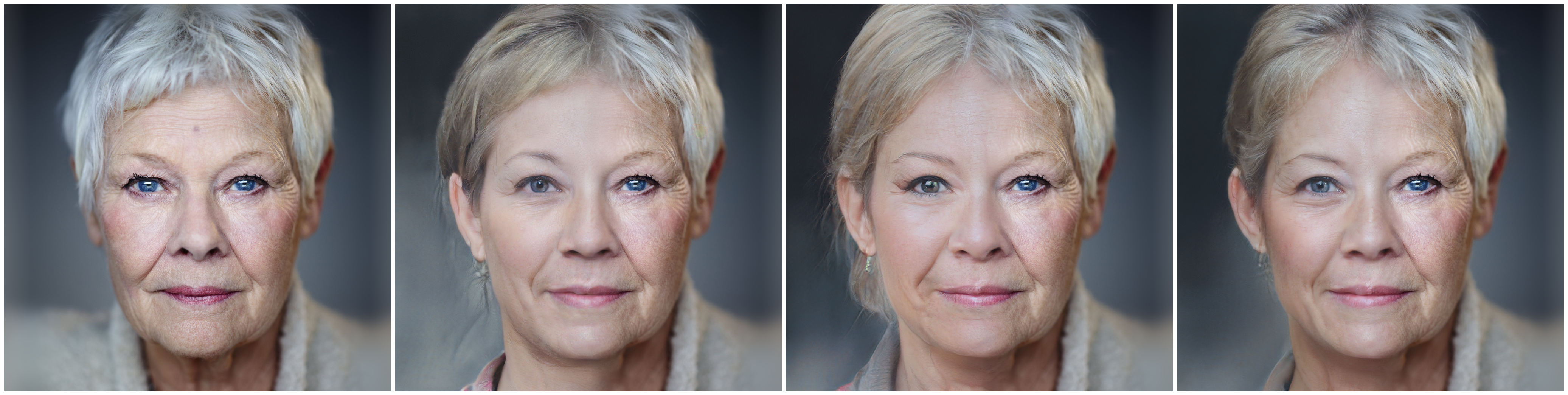}
        \includegraphics[width=\linewidth]{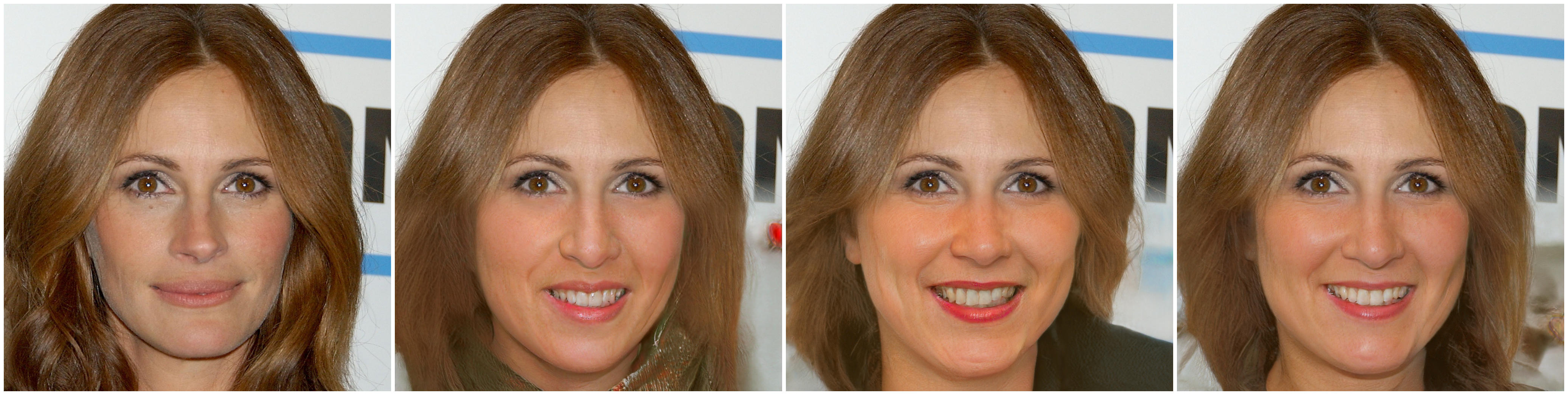}
        
        \caption{Inpainting results using different $w_{ini}$ initializations.}
        \label{fig:imp}
    \end{figure*}

In addition, our method is flexible and can generate different inpainting results (Fig.~\ref{fig:imp}), which cannot be fulfilled by any of the above-mentioned methods. All our inpainting results are of high perceptual quality.

\paragraph{Limitations}
Although better than the two state-of-the-art methods, our inpainting results still leave room for improvement.
For example in Fig.~\ref{fig:imp_1}, the lighting condition (first row), age (second row) and skin color (third and last row) are not learnt that well.
We propose to address them in the future work.

\small

 \begin{table*}[h]
     \centering
\begin{tabular}{lccc|ccc|ccc} \toprule
\multirow{2}{*}{Method}   & \multicolumn{3}{c|}{Image Resolution ($1024\times1024$)}  & \multicolumn{3}{c|}{Image Resolution ($512\times512$)}  & \multicolumn{3}{c}{Image Resolution ($256\times256$)}\\  
{}     & SSIM  & MSE  & PSNR           & SSIM  & MSE  & PSNR      & SSIM  & MSE  & PSNR \\ \midrule
                       
Partial Convolution~\cite{Liu_2018}  &0.8957    &199.39   &21.83         &0.8865    & 98.83  &21.92       &0.8789   &48.39  &22.17    \\
Gated Convolution~\cite{yu2018free}  &0.8693   &246.46   &19.65         &0.8568    &121.98   &19.77       &0.8295   &61.82  &19.41       \\
{\bf Ours} &{\bf 0.9176}&{\bf 180.69}&{\bf 22.35}  &{\bf 0.9104}& {\bf 89.25}& {\bf 22.48} &{\bf 0.9009}&{\bf 43.85}&{\bf 22.65}  \\
\bottomrule

\end{tabular}
     \caption{Evaluation results of image inpainting methods using SSIM, MSE and PSNR score.}
     \label{tab:tab1}
 \end{table*}

  \begin{table*}[t]
    \centering
    \begin{tabular}{ l r r }
    \toprule
        Edit & Attribute & Change in confidence \\ \hline
        \hline
        Wrinkle Smoothing  &   age & 21\% \\
        Adding Baldness &    bald & 75\% \\ 
        Adding Beard &    beard & 42\% \\
        Adding Moustache  & moustache & 49\% \\

    \bottomrule
    \end{tabular}
    \caption{Changes in confidence scores of classifier after user edits.}
    \label{tb:defect}
\end{table*}

\begin{figure*}
        \centering
        \includegraphics[width=\linewidth]{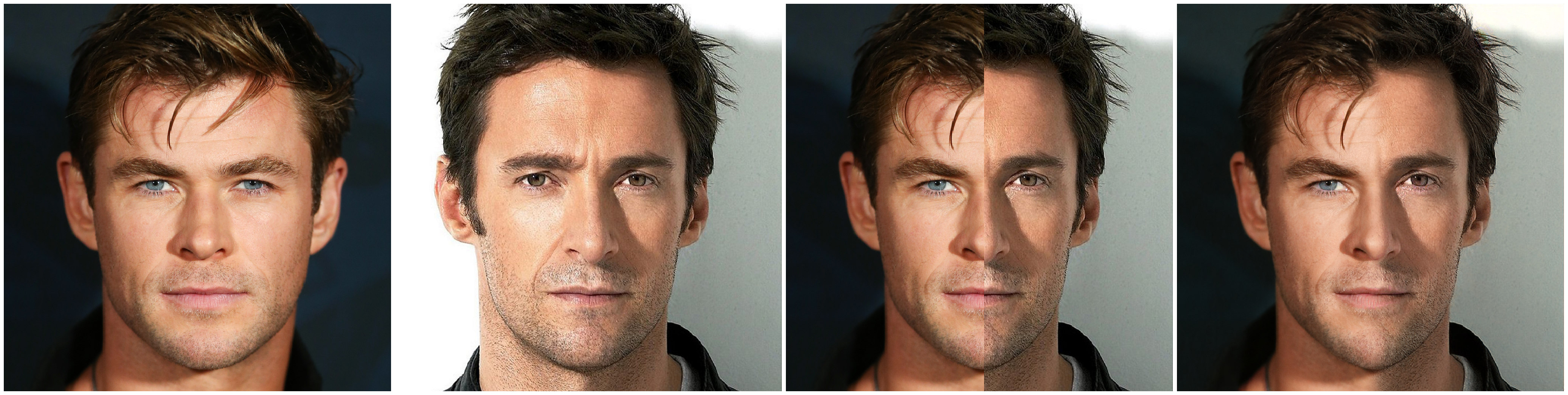}
        \includegraphics[width=\linewidth]{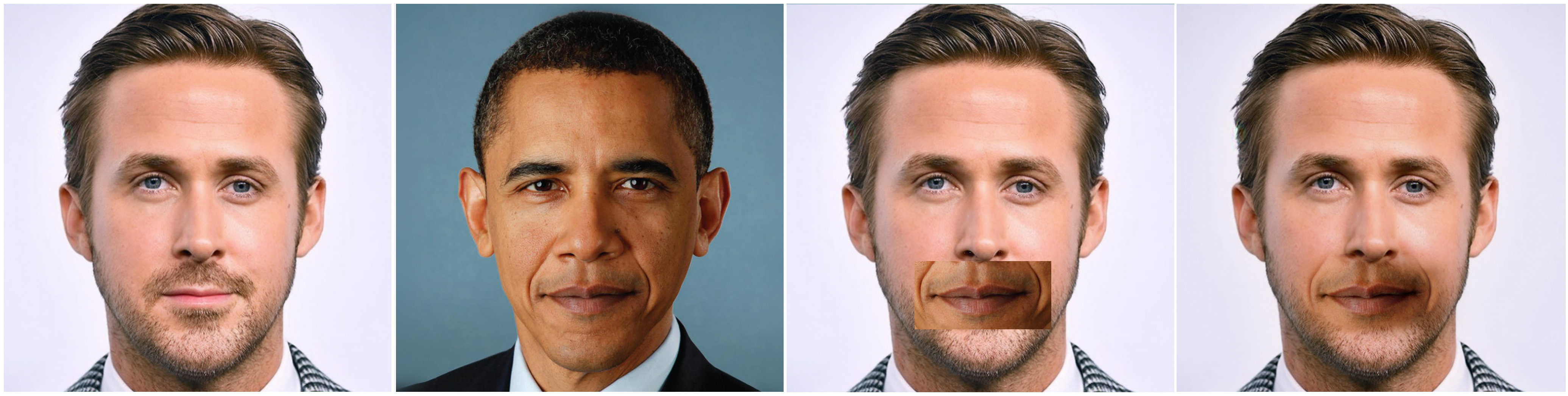}
        \includegraphics[width=\linewidth]{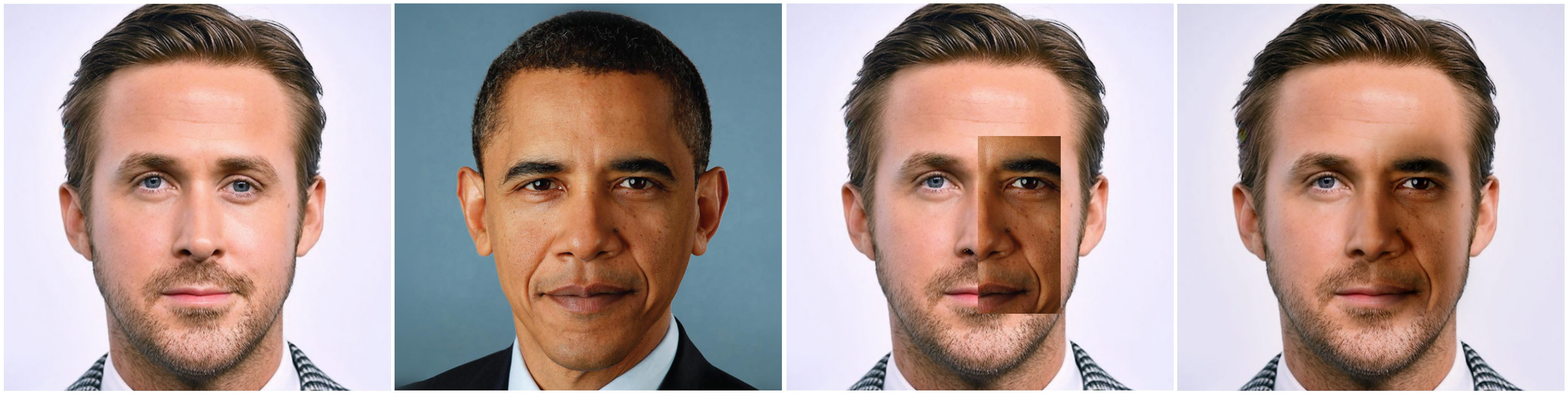}
        \caption{(a) and (b): input images; (c): the ``two-face'' generated by naively copying the left half from (a) and the right half from (b); (d): the ``two-face'' generated by our Image2StyleGAN++ framework.}
        \label{fig:imp_next}
    \end{figure*}

 \begin{table*}[t]
    \centering
    \begin{tabular}{ l l r r }
    \toprule
        Pretrained model & Interpolation & Perceptual length (full) & Perceptual length (end) \\ \hline
        \multirow{2}{*}{FFHQ} & Non-Masked & 227.1  &  191.1\\ 
                             & Masked & 112.1  &  89.8\\ \hline
        \multirow{2}{*}{LSUN Cars} & Non-Masked & 12388.1 & 6038.5 \\ 
                             & Masked & 4742.3 & 3057.9 \\\hline
        \multirow{2}{*}{LSUN Bedrooms} & Non-Masked & 2521.1 & 1268.7 \\ 
                             & Masked & 1629.8 & 938.1 \\
    
    \bottomrule
    \end{tabular}
    \caption{Perceptual length evaluation for masked and non-masked interpolation.}
    \label{tb:inter}
\end{table*}

\subsection{Image Crossover}
To further evaluate the expressibility of the Noise space, we show additional results on image crossover in Fig.~\ref{fig:imp_next}. We show that the space is able to crossover parts of images from different races (see second and third column).

\subsection{Local Edits using Scribbles}
\begin{figure*}[t]
        \centering
        \begin{subfigure}{0.49\textwidth}
            \includegraphics[width= 0.99\linewidth]{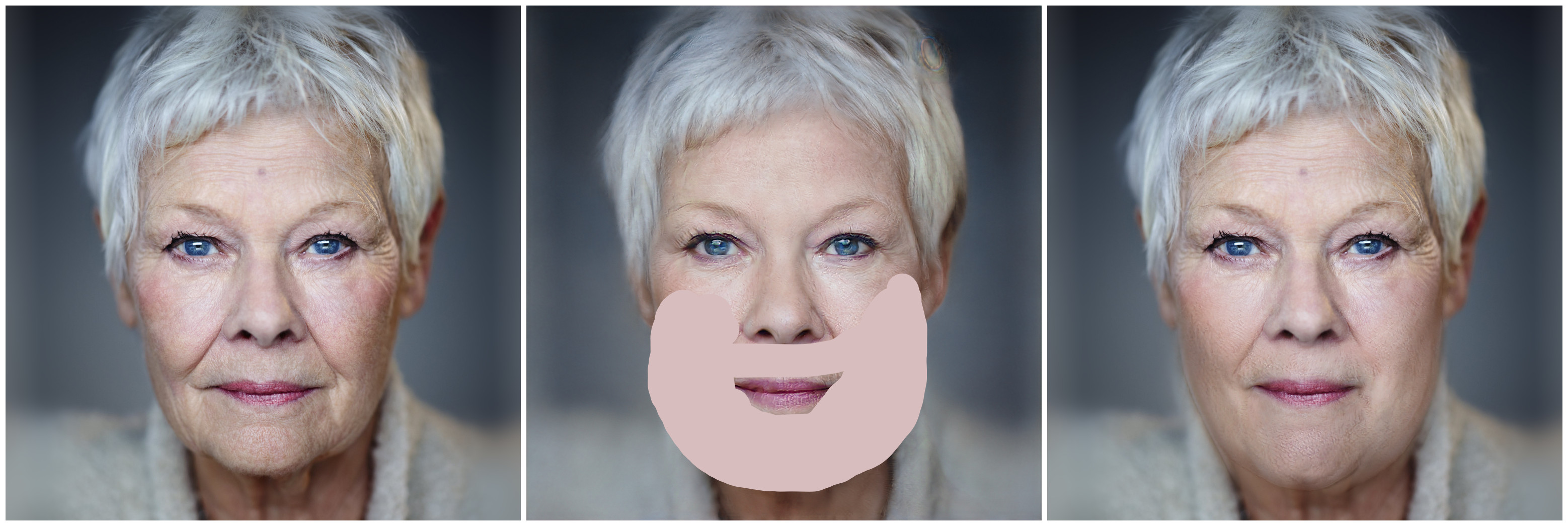}
             \includegraphics[width=0.99\linewidth]{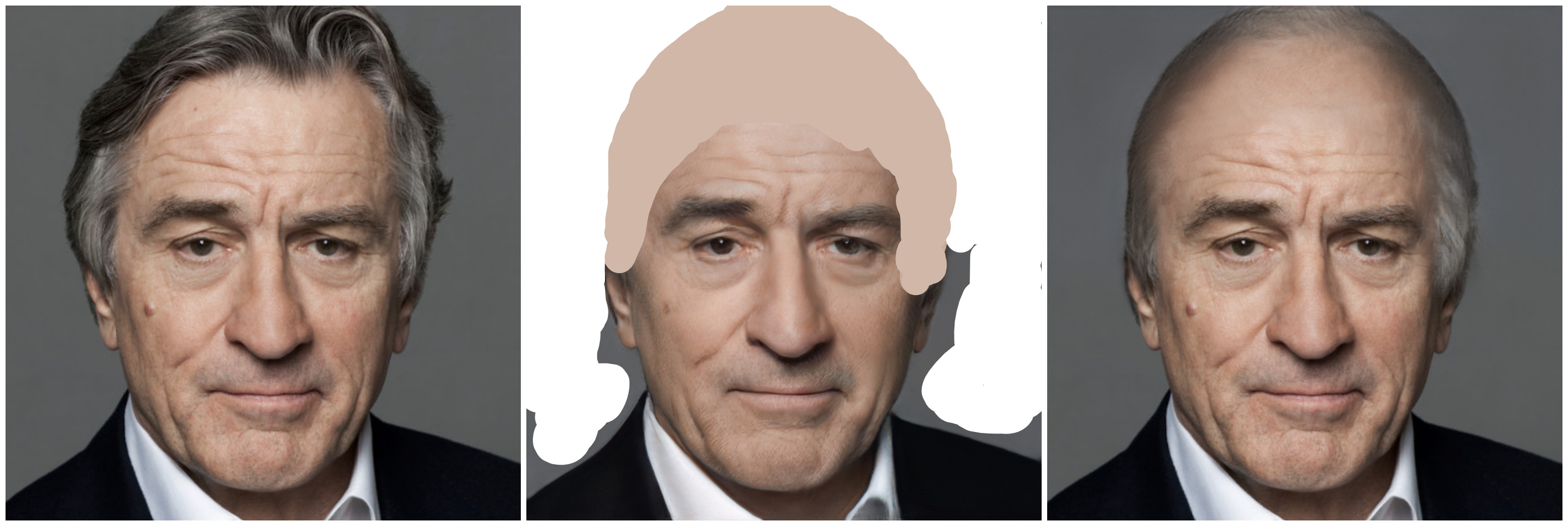}
             \includegraphics[width=0.99\linewidth]{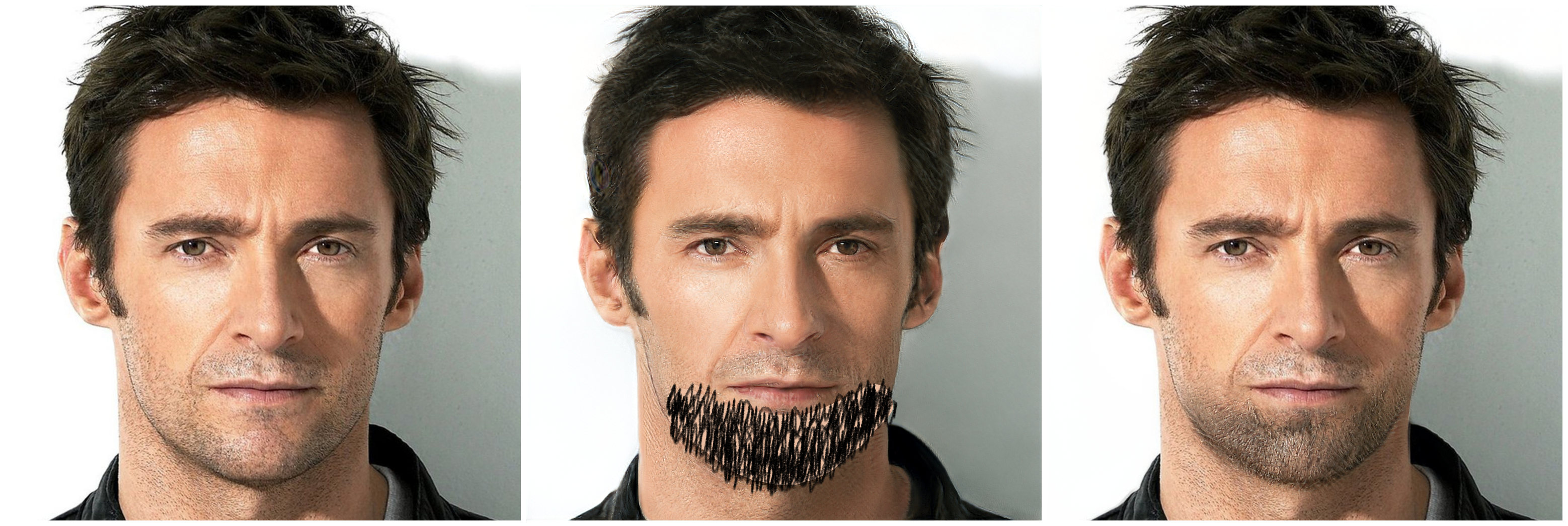}
             
        \end{subfigure}
        \begin{subfigure}{0.49\textwidth}
             \includegraphics[width=0.99\linewidth]{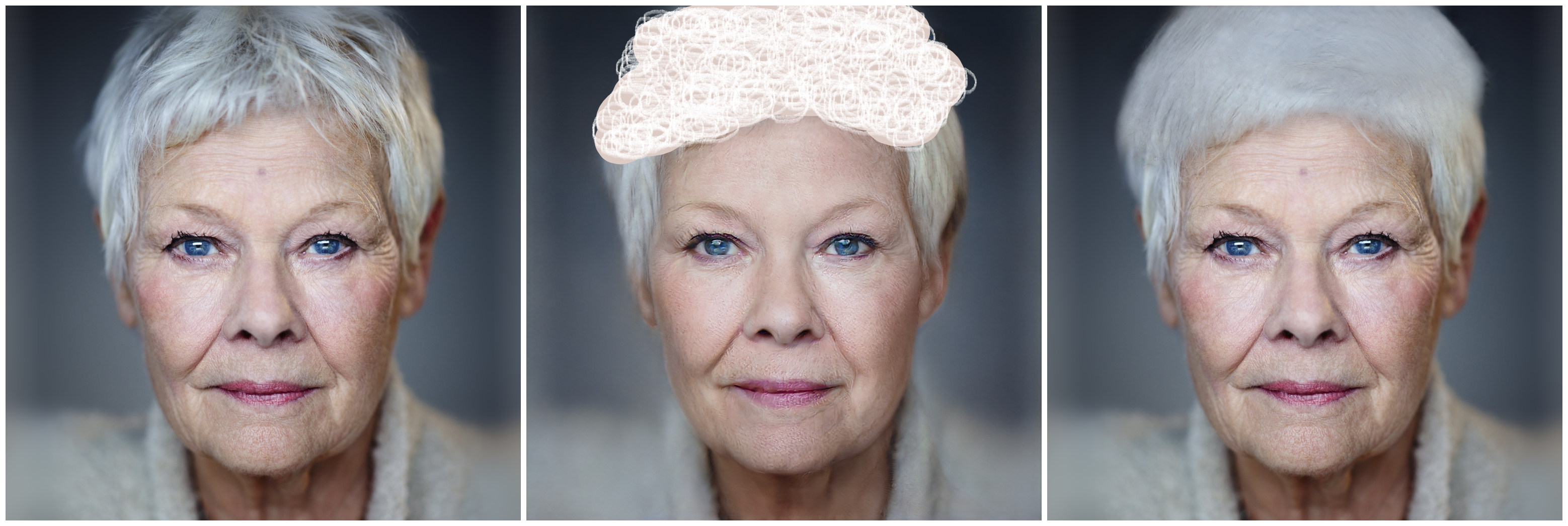}
              \includegraphics[width=0.99\linewidth]{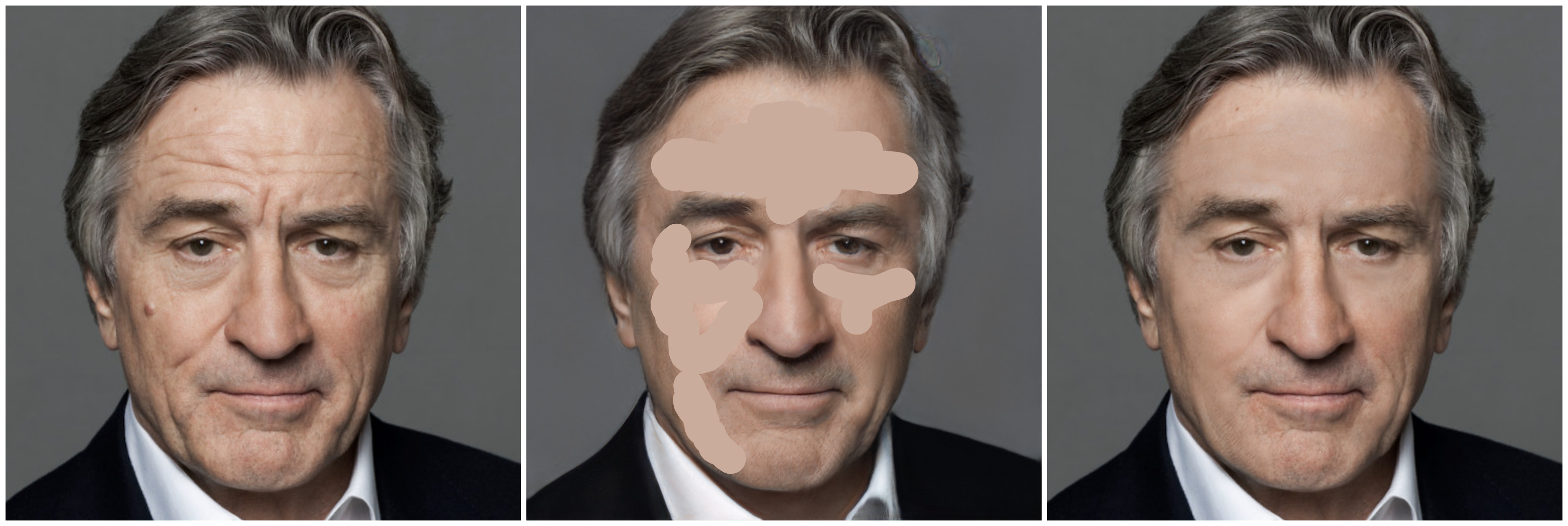}
               \includegraphics[width=0.99\linewidth]{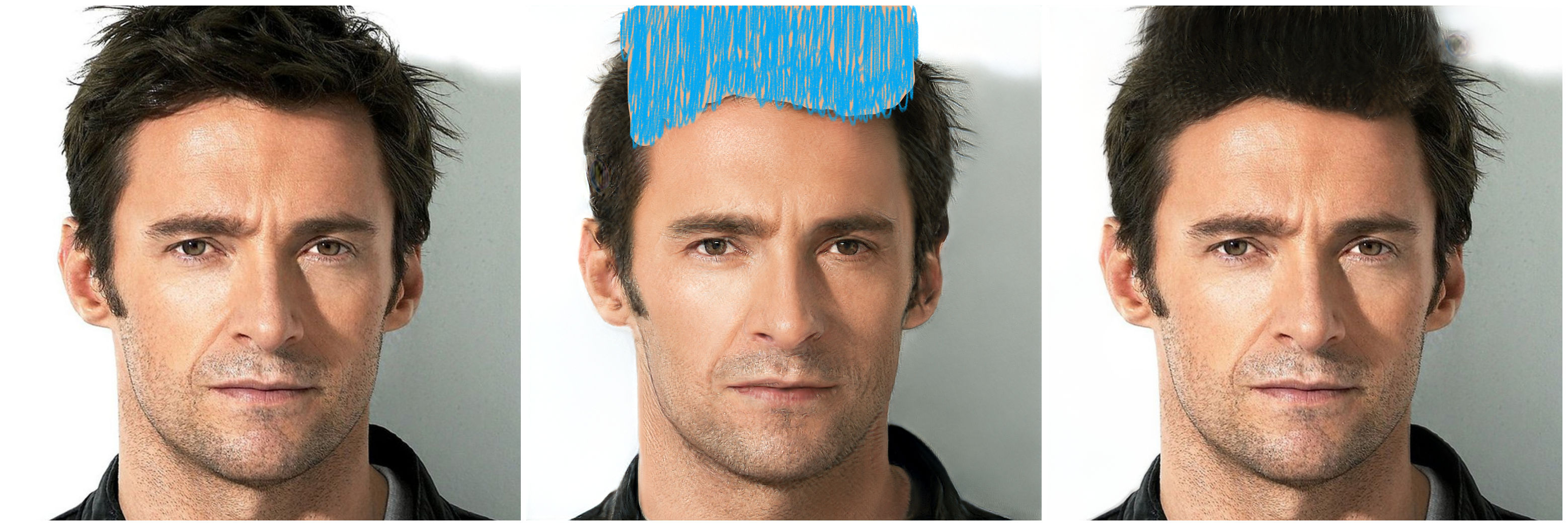}
        \end{subfigure}
        \caption{Column 1 \& 4: base image; Column 2 \& 5: scribbled image ; Column 3 \& 6: result of local edits.}
        \label{fig:edit}
    \end{figure*}

In order to evaluate the quality of the local edits using scribbles, we evaluate the face attribute scores~\cite{git2} on edited images. We perform some common edits of adding baldness, adding a beard, smoothing wrinkles and adding a moustache on the face images to evaluate how photo-realistic the edited images are. Table~\ref{tb:defect} shows the average change in the confidence of the classifier after a particular edit is performed. We also show additional results of the Local edits in Fig. \ref{fig:edit}. For our method, one remaining challenge is that sometimes the edited region is overly smooth (e.g. first row).

\subsection{Attribute Level Feature Transfer}

\begin{figure*}
        \centering
        \includegraphics[width=\linewidth]{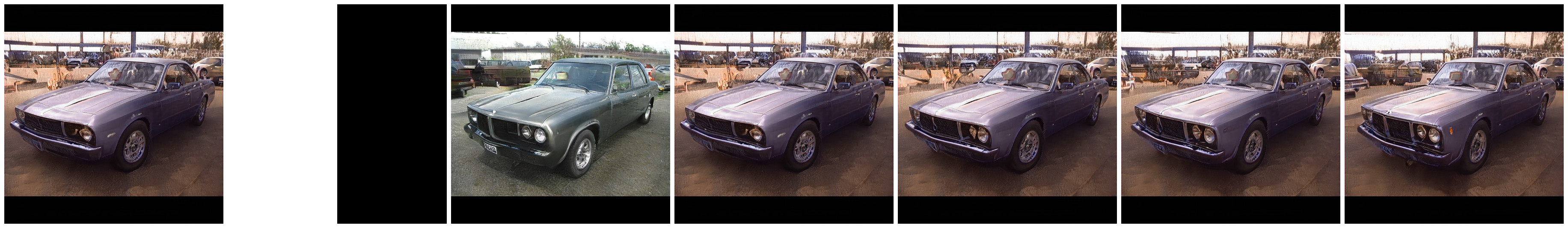}
        \includegraphics[width=\linewidth]{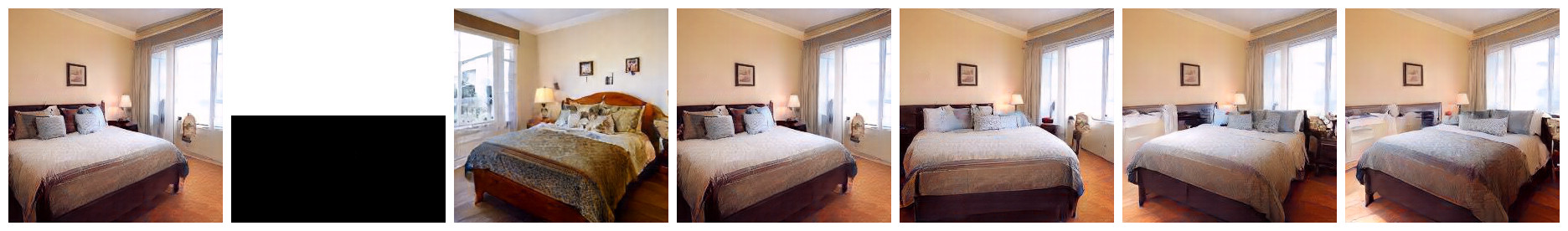}
        \caption{First column: base image; Second column: mask area; Third column: attribute image; Fourth to Eighth column: image generated via attribute level feature transfer and masked interpolation.}
        
        \label{fig:carbed}
    \end{figure*}

We show a video in which attribute interpolation can be performed on the base image by copying the content from an attribute image. Here different attributes can be taken from different images embedded in the $W^+$ space and applied to the base image. These attributes can be independently interpolated and the results show that the blending quality of the framework is quite high. We also show additional results on LSUN Cars and LSUN Bedrooms in the video (also see Fig.~\ref{fig:carbed}). Notice that in the LSUN bedrooms, for instance, the style and the position of the beds can be customized without changing the room layout. 

In order to evaluate the perceptual quality of attribute level feature transfer, we compute perceptual length~\cite{STYLEGAN2018} between the images produced by independently interpolated attributes (called masked interpolation). StyleGAN~\cite{STYLEGAN2018} showed that the metric evaluates how perceptually smooth the transitions are.  Here, perceptual length measures the changes produced by feature transfer which may be affected especially by the boundary of the blending. The boundary may tend to produce additional artifacts or introduce additional features which is clearly undesirable.

We compute the perceptual length across 1000 samples using two masks shown in Fig.~\ref{fig:masks} (First and Seventh column). In Table~\ref{tb:inter} we show the results of the computation of the perceptual length (both for masked and non-masked interpolation) on FFHQ, LSUN Cars and LSUN Bedrooms pretrained StyleGAN. We compare these scores as the non-masked interpolation gives us the upper bound of the perceptual length for a model (in this case there is no constraint on what features of the face should change). As a particular area of the image is interpolated rather than the whole image, note that our results on FFHQ pretrained StyleGAN produce lower score than the non-masked interpolation. The low perceptual length score suggests that there is a less drastic change. Hence, we conclude that the output images have comparable perceptual quality with non-masked interpolation. 

LSUN Cars and LSUN Bedrooms produce relatively higher perceptual length score. We attribute this result to the fact that the images in these datasets can translate and the position of the features is not fixed. Hence, the two images produced at random might have different orientation in which case the blending does not work as good.

 \subsection{Channel wise feature average}
\begin{figure*}[h]
          \centering
       \includegraphics[width=\linewidth]{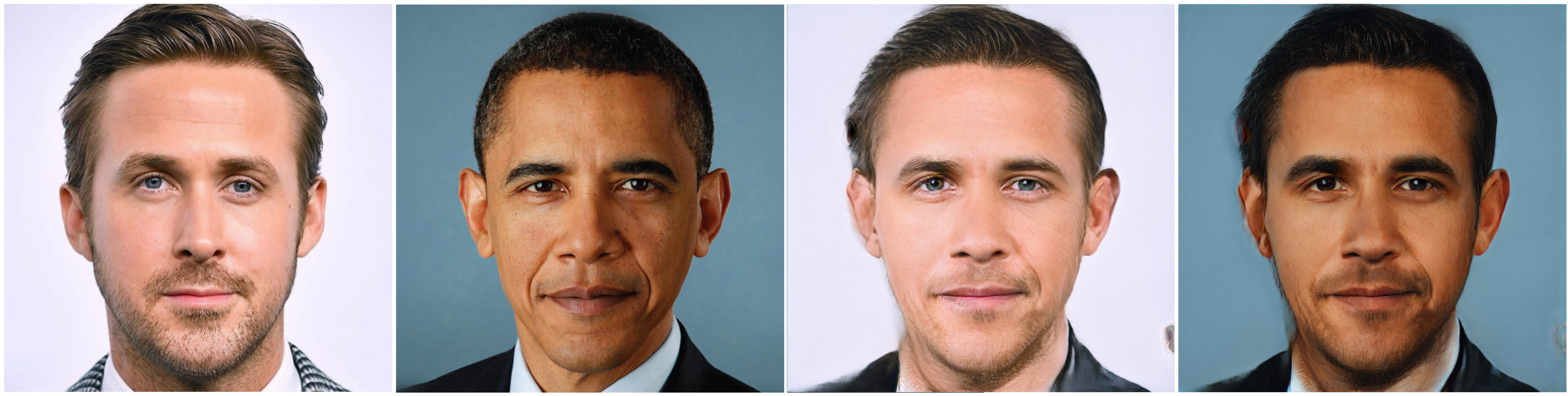}
      \includegraphics[width=\linewidth]{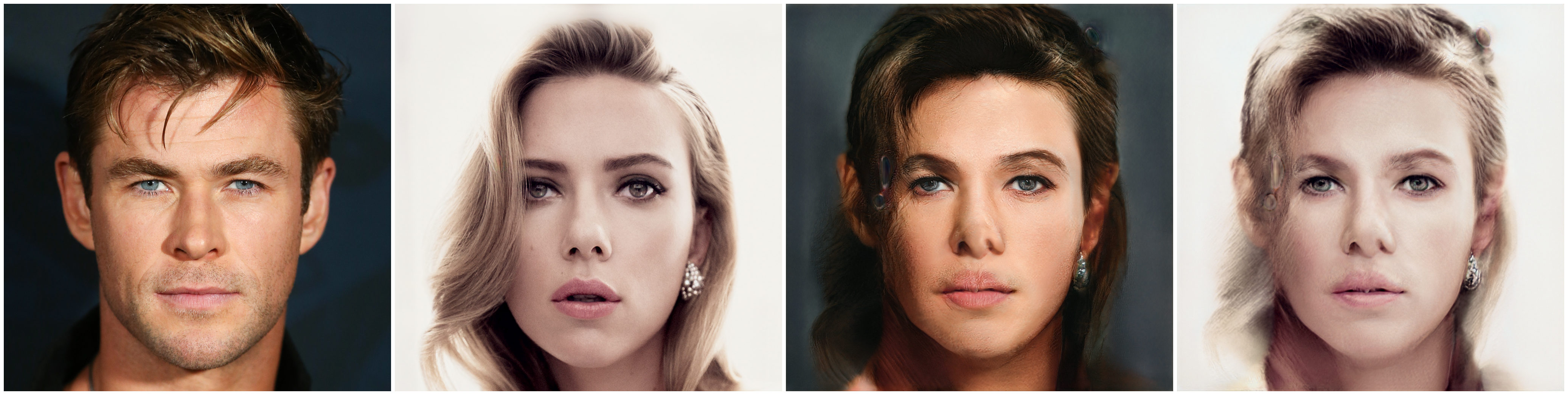}

          \caption{First column: First Image; Second Column: Second Image; Third Column: Feature averaged image  using $W^{+}$ code of first image; Fourth Column: Feature averaged image  using $W^{+}$ code of second image. }
          \label{fig:weee}
     \end{figure*}

 We perform another operation denoted by $I_{att}(1,0,w_{x},,n_{ini}, 6 )$,  where $w_{x}$ can be the $W^{+}$ code for images $I_{1}$ or $I_{2}$. In Fig.~\ref{fig:weee}, we show the result of this operation which is initialized with two different $W^{+}$ codes. The resulting faces contain the characteristics of both faces and the styles are modulated by the input $W^{+}$ codes.
 
\end{document}